\documentclass[11pt]{article}

\usepackage[a4paper,margin=1in]{geometry}

\usepackage{cite}
\usepackage{amsmath,amssymb,amsfonts}
\usepackage{algorithm}
\usepackage{algorithmic}
\usepackage{graphicx}
\usepackage{multirow}
\usepackage{manyfoot}
\usepackage{booktabs}
\usepackage{longtable}
\usepackage{makecell}
\usepackage{textcomp}
\usepackage{array}

\usepackage[hidelinks]{hyperref}

\def\BibTeX{{\rm B\kern-.05em{\sc i\kern-.025em b}\kern-.08em
    T\kern-.1667em\lower.7ex\hbox{E}\kern-.125emX}}    

\begin{document}

\title{Using Machine Learning to Investigate Predictors of Fasting Blood Glucose: Insights into Circadian Timing and Age Interactions}

\author{
V. Bu-Dager \qquad S. Cirstea\\[0.5em]
\small School of Computing and Information Science, Anglia Ruskin University\\
\small East Road, Cambridge CB1 1PT, U.K.\\
\small \texttt{vb37@aru.ac.uk} \qquad
\texttt{silvia.cirstea@aru.ac.uk}
}

\maketitle

\begin{abstract} Impaired glucose regulation is a major contributor to metabolic dysfunction and type 2 diabetes. This study developed an interpretable machine-learning framework to predict log-transformed fasting blood glucose using metabolic, hormonal, lifestyle, demographic, nutritional, and circadian variables from the National Health and Nutrition Examination Survey 2017--2020 pre-pandemic dataset. After merging multiple NHANES sub-datasets, data processing used a leakage-resistant pipeline in which imputation, scaling, and one-hot encoding were performed only after dataset splitting and within training folds. Elastic Net, LASSO, and XGBoost models were evaluated using 94 candidate predictors and engineered circadian interaction terms. Performance was assessed using mean absolute error, root mean squared error, coefficient of determination, calibration, and Shapley Additive Explanations. The final interaction-augmented XGBoost model achieved strong performance on the independent test set, with a mean absolute error of 0.0804, a root mean squared error of 0.1148, and a coefficient of determination of 0.7761, using 10 predictors. Glycohemoglobin was the dominant predictor, followed by insulin, diabetes diagnosis, gamma-glutamyl transferase, age, race, and gender. Among the engineered interaction terms, sleep midpoint multiplied by age was consistently retained in repeated random-split analyses, although its contribution remained modest relative to dominant glycaemic predictors. These findings support further investigation of circadian-age interactions in metabolic health. \end{abstract} 


\vspace{1em}

\section{Introduction}\label{sec:introduction}

Metabolic health is an important contributor to overall well-being and better quality of life. Metabolic regulation involves intricate processes and pathways to maintain cellular homeostasis. A key metabolic regulator is insulin, a hormone that controls blood glucose levels by promoting glucose metabolism \cite{nortonInsulinMasterRegulator2022}. Metabolic dysfunction leads to several disease manifestations such as diabetes, which is associated with long-term complications including renal failure, cardiovascular disease, and critical nerve damage. There are two main types of diabetes, type 1 diabetes (T1D) and type 2 diabetes (T2D).  T1D is an autoimmune condition in which insulin-producing pancreatic cells are destroyed, whereas in T2D, insulin resistance is a key factor due to a lack of cellular insulin response or inadequate insulin production \cite{krauseType1Type2023}. Diabetes is commonly identified through abnormal blood glucose measures, including fasting plasma glucose and glycohemoglobin (HbA1c; labelled as GHB in figures and model outputs) \cite{zhangEarlyDetectionType2023}. Both fasting glucose and HbA1c are widely used indicators of glycaemic status and glucose dysregulation.

Multiple established factors contribute to the development of T2D, including obesity, family history, poor diet, physical inactivity, and age \cite{wuRiskFactorsContributing2014}. Some studies have shown that circadian rhythms can also influence glucose regulation through central and peripheral clocks synchronised by environmental cues such as light exposure, feeding, and activity cycles \cite{manoogianCircadianClockNutrient2016, zhangCircadianRhythmGlucose2025}. In fact, disruption of circadian rhythm, which is commonly observed in shift workers or irregular sleepers, has been linked to impaired glucose tolerance and reduced insulin sensitivity \cite{morrisEffectsInternalCircadian2016, opperhuizenLightNightAcutely2017, tranEffectCircadianClock2024}. Moreover, personalized circadian glucose profiles suggest that sleep midpoint timing may be relevant to glycaemic regulation independent of diet or activity \cite{phillipsUncoveringPersonalizedGlucose2023}. Sleep midpoint, defined as the halfway point between sleep onset and offset, is often used as a marker of circadian alignment and reflects individual chronotype and sleep--wake timing, both of which have been associated with metabolic processes \cite{yuEveningChronotypeAssociated2015, maghsoudipourAssociationsChronotypeSleep2022}. 

To our knowledge, prior machine-learning studies of diabetes and glycemic prediction have only rarely incorporated sleep-timing or circadian-related variables, and we found no study that systematically combines such variables with engineered interaction terms for population-based fasting blood glucose prediction. Existing ML studies in diabetes and related metabolic risk prediction have primarily focused on established clinical and demographic features such as body mass index (BMI), age, triglycerides, hypertension, and fasting glucose-related markers \cite{liuMachineLearningModels2023, laiPredictiveModelsDiabetes2019, liMachineLearningPredicting2023, lvDetectionDiabeticPatients2023, taoPredictingThreemonthFasting2023, elmagarmidInvestigationRiskFactors2024, hossainMetabolicSyndromePredictive2024, shojaee-mendPredictionDiabetesUsing2024}. While some more recent studies have begun to incorporate broader behavioural, physiological, or time-structured inputs, such as physical activity monitoring, sleep disturbance, night work, or chronobiologically informed glucose features, these approaches remain limited and are typically focused on continuous glucose monitoring, incident diabetes, or non-population-based settings rather than fasting blood glucose prediction in a general adult cohort \cite{vandoornMachineLearningbasedGlucose2021, liuUseMachineLearning2024, burksChronobiologicallyinformedFeaturesCGM2025, kellerRiskDiabetesLong2025, kiranType2Diabetes2026}. Ensemble models such as XGBoost and stacking approaches have achieved strong predictive performance \cite{changApplicationMachineLearning2022, fuStackingModelFramework2024, zhangDevelopmentValidationMachine2024}, but the integration of circadian timing variables with explicit interaction modelling remains underexplored. 

This study aimed to develop an interpretable machine learning framework for prediction of fasting blood glucose, an important marker of glucose dysregulation, in a large, diverse adult cohort from the National Health and Nutrition Examination Survey (NHANES). To do so, we integrated a broad feature space including metabolic markers, hormonal indicators, circadian-related variables, lifestyle factors, nutritional variables, and demographic characteristics. In addition to evaluating established clinical predictors, we examined whether engineered circadian-demographic interaction terms could contribute useful predictive information within a multivariable framework. By combining interpretable machine learning with explicit modelling of circadian-age interactions, this study aimed to provide a more nuanced view of fasting glucose prediction and to explore whether circadian-related signals can be detected alongside dominant metabolic predictors.

\section{Materials and Methods}\label{sec:materials_and_methods}

\begin{figure*}[!t]
\centerline{\includegraphics[width=\textwidth]{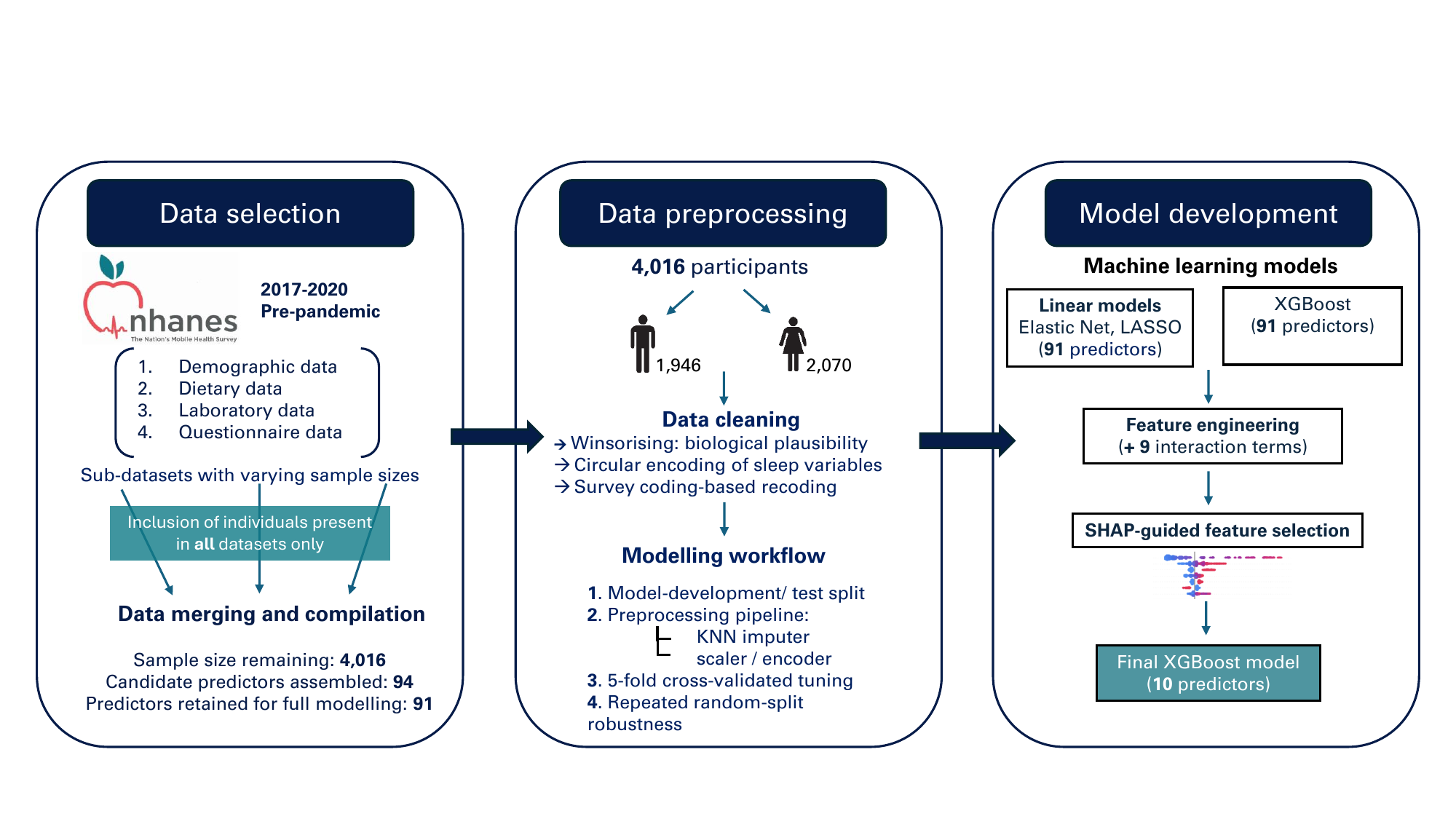}}
\caption{Overview of the study workflow. NHANES 2017--2020 pre-pandemic sub-datasets were merged to form the analytic sample ($n = 4{,}016$). A total of 94 candidate predictors were assembled across demographic, metabolic, circadian, lifestyle, hormonal, and nutritional domains. Preprocessing included biologically guided winsorization, circular encoding of sleep-related variables, survey coding-based recoding, and correlation-based filtering, after which 91 predictors remained for full-model development. The dataset was then split into a model-development set and an independent test set, with nested cross-validation applied within the development subset. Elastic Net. LASSO, and XGBoost models were trained, followed by interaction-based feature engineering, SHAP-guided feature selection, and construction of the final parsimonious interaction-augmented XGBoost model.}\label{fig:workflow}
\end{figure*}

The overall study design and analytical workflow are summarized in Fig.~\ref{fig:workflow}, including data selection, preprocessing, feature engineering, model development, and robustness assessment.

\subsection{Study Population}\label{subsec:study_population}

This study used data from the National Health and Nutrition Examination Survey (NHANES), focusing on the pre-pandemic 2017--2020 cycles. NHANES is a nationally representative survey conducted by the Centers for Disease Control and Prevention (CDC), combining interviews, physical examinations, and laboratory testing to assess the health and nutritional status of the civilian, non-institutionalized U.S. population. All data used in this study were publicly available and de-identified; therefore, no additional ethical approval was required. 

The analytic dataset was constructed by merging multiple NHANES sub-datasets with varying participant counts, ranging from 5,090 individuals with complete fasting glucose and insulin measurements to 15,560 individuals with demographic data. Datasets were merged using the participant identifier (SEQN). Prior to the final merge, 145 participants from the merged biomarker dataset (2.8\% of the initial 5,090) were excluded because more than 50\% of relevant variables were missing. Within the sleep dataset, 222 of 10,195 rows (2.2\%) were excluded because of missing values in critical variables required for derivation of circadian measures, specifically weekday/weekend bedtimes, wake times, and sleep duration. Participants with missing fasting glucose were also excluded because fasting glucose was the modelling target. The remaining missing predictor values were handled within the machine-learning pipeline using imputation. These filtering steps were applied before construction of the final analytic dataset to ensure adequate data quality and reliable derivation of circadian measures. The main sample-construction steps are summarized in Appendix Table~\ref{tab:sample_flow}. The final analytic sample included 4,016 adults aged 18 to 80 years, comprising 1,946 males and 2,070 females.

\subsection{Variable Selection and Domain Structure}\label{subsec:variable_selection}

To investigate factors associated with fasting blood glucose regulation, a broad set of variables was extracted from NHANES and grouped into six conceptual domains: metabolic biomarkers, circadian-related variables, lifestyle and medical history, hormonal profiles, nutritional intake, and control variables (Table~\ref{tab:variables}). Age, gender, race, and ratio of family income to poverty were included as control variables because of their established associations with metabolic dysregulation. This domain-based structure was designed to capture both direct metabolic processes and modifiable behaviours, such as sleep, physical activity, and diet, that may interact with circadian rhythms and influence glucose homeostasis. These domains were used to organize candidate predictors conceptually before correlation-based filtering and model-specific feature selection.

\begin{table}[t]
\caption{Variables included from the NHANES 2017--2020 pre-pandemic dataset.}
\label{tab:variables}
\centering
\begin{tabular}{p{0.23\columnwidth} p{0.70\columnwidth}}
\toprule
\textbf{Category} & \textbf{Variables} \\
\midrule
\emph{Metabolic biomarkers} &
Body mass index (BMI), waist circumference, high-density lipoprotein cholesterol (HDL), low-density lipoprotein cholesterol (LDL), glycohemoglobin (HbA1c; denoted as GHB in model variables), high-sensitivity C-reactive protein (CRP), alanine aminotransferase (ALT), aspartate aminotransferase (AST), gamma-glutamyl transferase (GGT), uric acid, triglycerides, estimated glomerular filtration rate (eGFR), insulin, total bilirubin, systolic blood pressure, diastolic blood pressure, fasting glucose, blood urea nitrogen (BUN), lactate dehydrogenase (LDH), globulin, total protein, total cholesterol, and creatinine \\
\midrule

\emph{Circadian-related variables} &
Bedtime, sleep health index, social jetlag hours, sleep apnoea risk, sleep midpoint (weekdays and weekends), sleep schedule category, circadian disruption indicators, and sleep duration variability \\
\midrule

\emph{Lifestyle and medical history} &
Physical activity levels (work-related and recreational), family history of diabetes or cancer, smoking status (smoked at least 100 cigarettes in lifetime), sedentary time, lifestyle scores and categories derived from vigorous and moderate work/recreational activity and walking or bicycling, and doctor-diagnosed diabetes \\
\midrule

\emph{Hormonal profiles} &
Sex hormone-binding globulin (SHBG), estradiol, androstenedione, luteinizing hormone (LH), and follicle-stimulating hormone (FSH) \\
\midrule

\emph{Nutritional intake} &
Macronutrients (energy, carbohydrates, fat, fibre, and protein), micronutrients (vitamins and minerals), caffeine, alcohol, and dietary pattern indicators (e.g., sugar-free, low-carbohydrate, and high-protein diets) \\
\midrule

\emph{Control variables} &
Gender, age, race, and family income-to-poverty ratio \\
\bottomrule
\end{tabular}
\end{table}

\subsection{Variable Transformation and Feature Engineering}\label{subsec:feature_engineering}

Logarithmic transformation was applied to skewed biochemical markers, including fasting glucose, insulin, high-sensitivity C-reactive protein (CRP), triglycerides, alanine aminotransferase (ALT), aspartate aminotransferase (AST), and gamma-glutamyl transferase (GGT), to improve distributional symmetry and model convergence.

To capture the periodic nature of sleep and wake times on a 24-hour clock, all relevant time variables (e.g., weekday and weekend bedtimes and wake times) were circularly encoded using sine and cosine transformations:
\begin{equation}
    \sin_t = \sin\left(2\pi \frac{t}{24}\right)
\end{equation}
\begin{equation}
    \cos_t = \cos\left(2\pi \frac{t}{24}\right)
\end{equation}
where $t$ denotes time in decimal hours (e.g., 23:30 = 23.5). This encoding preserves the proximity of times across the midnight boundary (e.g., 23:00 and 1:00).

Additional indices of circadian misalignment and sleep regularity were then derived:
\begin{itemize}
    \item \textbf{Social Jetlag:} Calculated as the angular distance (in hours) between weekday and weekend bedtimes after circular encoding, thereby reflecting the true timing shift independently of clock time:
    \begin{equation}
    \begin{aligned}
    J_{\mathrm{s}}
    &=
    2 \arcsin\!\left(
    \frac{1}{2}
    \sqrt{
    (\sin_w - \sin_{wd})^2 + (\cos_w - \cos_{wd})^2
    }
    \right)\\
    &\quad \times \frac{24}{2\pi},
    \end{aligned}
    \end{equation}
    where $J_{\mathrm{s}}$ denotes social jetlag in hours, $\sin_{wd}$, $\cos_{wd}$, $\sin_w$, and $\cos_w$ denote the sine and cosine encodings of weekday and weekend bedtimes, respectively.

    \item \textbf{Sleep Midpoint:} Calculated as bedtime plus half of total sleep duration, modulo 24.

    \item \textbf{Sleep Regularity:} Assessed as the standard deviation of sleep timing and duration across weekdays and weekends.

    \item \textbf{Sleep Health Index:} A composite sleep score calculated from weekday and weekend sleep duration, with penalties for greater jetlag and self-reported sleep problems; higher values indicate more favourable sleep patterns.
\end{itemize}

The Python code used to derive engineered sleep and circadian-disruption variables is provided in Appendix Fig.~\ref{fig:supp_code_sleep}. 

A total of nine interaction terms were engineered a priori to capture potential interplay between circadian, metabolic, and demographic factors, including Average Sleep Hours $\times$ Insulin, Sleep Midpoint $\times$ Age, Circadian Disruption $\times$ Age, Social Jetlag $\times$ Age, Sleep Duration $\times$ Age, Bedtime $\times$ Age, Circadian Disruption $\times$ GHB, Waist Circumference $\times$ Jetlag, and Sleep Midpoint $\times$ Insulin. These interaction terms were designed to reflect biologically plausible joint effects on fasting glucose regulation. Age-modulated interaction terms were included because prior studies suggest that the metabolic consequences of circadian disruption may vary across the life course. The hierarchical correlation matrix (Appendix Fig.~\ref{fig:cluster_heatmap}) showed clustering between metabolic, hormonal, demographic, and selected circadian features, providing additional motivation for exploring cross-domain interactions. All nine candidate interaction terms were entered together into the interaction-augmented XGBoost model (Model 4), and their contribution was subsequently assessed using SHAP-based feature selection and repeated random-split robustness analyses.

\subsection{Data Preprocessing and Dataset Splitting}\label{subsec:preprocessing}

All merged data were cleaned and checked for plausibility before analysis. Sleep-related time variables were transformed using circular encoding to preserve their temporal structure. Implausible or extreme values in physiological and biochemical variables were winsorized to clinically established, biologically plausible ranges (e.g., BMI: 15--50\,kg/m$^2$, HbA1c (denoted as GHB in model variables): 3--12\%, fasting glucose: 50--300\,mg/dL, insulin: 1--100\,$\mu$IU/mL) prior to modelling. These knowledge-based bounds, defined using external clinical guidance rather than the empirical study distribution, were applied to control outliers without introducing data leakage.

To reduce multicollinearity and improve model efficiency, highly correlated predictors (Pearson correlation coefficient $>$ 0.9) were removed. The resulting dataset was randomly split into a model-development set (80\%) and an independent test set (20\%), stratified by age group. The test set included 219 young adults (18--35 years), 323 middle-aged adults (35--60 years), and 262 older adults (60+ years), enabling age-specific evaluation of model performance.

All data-driven preprocessing steps, including missing-value imputation, feature scaling, and one-hot encoding of categorical variables, were performed only within the training portion of each resampling step using a \texttt{scikit-learn} pipeline. Specifically, missing values were imputed using K-nearest neighbours (KNN) imputation ($k = 10$), continuous variables were standardized, and categorical variables were one-hot encoded. Categorical-variable definitions were specified before model training through survey coding-based recoding and explicit variable typing (e.g., assigning an explicit ``unknown'' category to ambiguous responses), whereas the one-hot encoder itself was fit only within the training portion of each resampling step and then applied to the corresponding validation or test data using the same \texttt{ColumnTransformer} pipeline. Only survey coding-based recoding and the above knowledge-driven winsorization were performed before splitting, as these steps did not rely on the observed data distribution. This design prevented information leakage from held-out data during model development.

NHANES survey weights, stratification, and clustering variables were not incorporated into the machine learning models. This study was designed as a predictive modelling analysis within a harmonized analytic dataset rather than a survey-weighted population inference study. Because multiple NHANES components with differing subsample structures were merged and evaluated using repeated data splitting and cross-validated pipelines, the findings should be interpreted as internally validated predictive results for the assembled sample rather than nationally representative weighted estimates.

\subsection{Model Development and Evaluation}\label{subsec:model_development_evaluation}

Seven model variants were evaluated for prediction of log-transformed fasting glucose. Elastic Net and LASSO regression were used as interpretable linear benchmark models, whereas XGBoost models were used to capture nonlinear relationships and higher-order interactions. Hyperparameter tuning was performed within the model-development subset using five-fold cross-validated \texttt{RandomizedSearchCV}, with 20 iterations for Elastic Net and LASSO and 50 iterations for XGBoost, optimizing the $R^2$ metric. Full hyperparameter search spaces and optimal parameter settings for each model are reported in Appendix~\ref{secA4} (Tables~\ref{tab:hyperparameter_space} and~\ref{tab:optimal_hyperparameters}).

\subsubsection{Model development}

Model 1 was an Elastic Net regression model using the full feature set. Model 1b was a LASSO regression model using the same full feature set. Model 2 was an XGBoost model trained on the full feature set. Model 3 was an XGBoost model trained on a SHAP-guided reduced predictor set. Model 4 extended the full XGBoost model by including the engineered interaction terms described above. Model 5 combined interaction-based modelling with reduced-feature selection, and subsequent pruning yielded the final Model 5b. The final model was selected based on its balance of predictive performance, robustness across repeated random splits, and interpretability.

\subsubsection{SHAP-guided feature selection}

Reduced-feature XGBoost models were derived using SHapley Additive exPlanations (SHAP)-guided feature selection estimated exclusively within the model-development data. Specifically, SHAP values from the full XGBoost model (Model 2) were aggregated to the raw predictor level and used to rank candidate features. Model 3 was then trained using the top-ranked reduced feature set. An analogous procedure was applied to the interaction-augmented XGBoost model, yielding Model 5 from Model 4. In this interaction-augmented branch, SHAP-guided reduction was applied directly to Model 4, such that the original predictors and engineered interaction terms were ranked together within a single combined selection procedure. Model 5b further pruned this reduced interaction-augmented model by removing features with negative permutation importance in the evaluation pipeline, resulting in the final parsimonious XGBoost model. Test-set SHAP values were used only for final interpretability analyses and were not used for feature selection.

\subsubsection{Performance assessment and robustness evaluation}

Model performance was assessed using complementary metrics reflecting predictive accuracy, generalization, robustness, calibration, and interpretability. Mean absolute error (MAE), root mean squared error (RMSE), and test-set $R^2$ were used to evaluate predictive performance on unseen data. Five-fold cross-validated $R^2$ within the model-development subset was used to summarize tuning performance. Model calibration was assessed using calibration slope and intercept estimated from a linear regression of observed on predicted values in the independent test set. 

Repeated random-split robustness analyses across 10 random seeds were conducted for the interaction-augmented XGBoost models to assess the stability of both predictive performance and SHAP-guided feature selection. For each seed, the same age-stratified train-test split, tuning, SHAP-guided reduction, pruning, and evaluation pipeline was repeated. In addition, restricted sensitivity analyses were performed after excluding major glycaemic predictors and related interaction terms to examine the extent to which predictive performance and circadian-feature selection depended on these dominant variables. Model interpretability was examined using SHAP, a widely used framework for improving the transparency of machine learning models \cite{huangIncreasingTransparencyMachine2023}. SHAP values were computed on held-out predictions using the corresponding model-development data as background. SHAP summaries were used to quantify global feature importance and generate local explanations.

\section{Results}\label{sec:results}

\subsection{Distribution of diabetic and sleep-related features across the study population}\label{subsec:distribution_diabetic_sleep_related_features}

A broad set of features spanning metabolic, circadian, lifestyle, hormonal, and nutritional domains was analysed to characterize the study population and assess patterns relevant to fasting glucose regulation. The model-development and independent test sets showed broadly similar distributions across variables. To contextualize the prediction task, fasting blood glucose (FBG) and HbA1c, two widely used indicators of glycaemic status, were examined visually using violin and scatter plots.

\begin{figure}[!t]
\centering
\includegraphics[width=0.9\columnwidth]{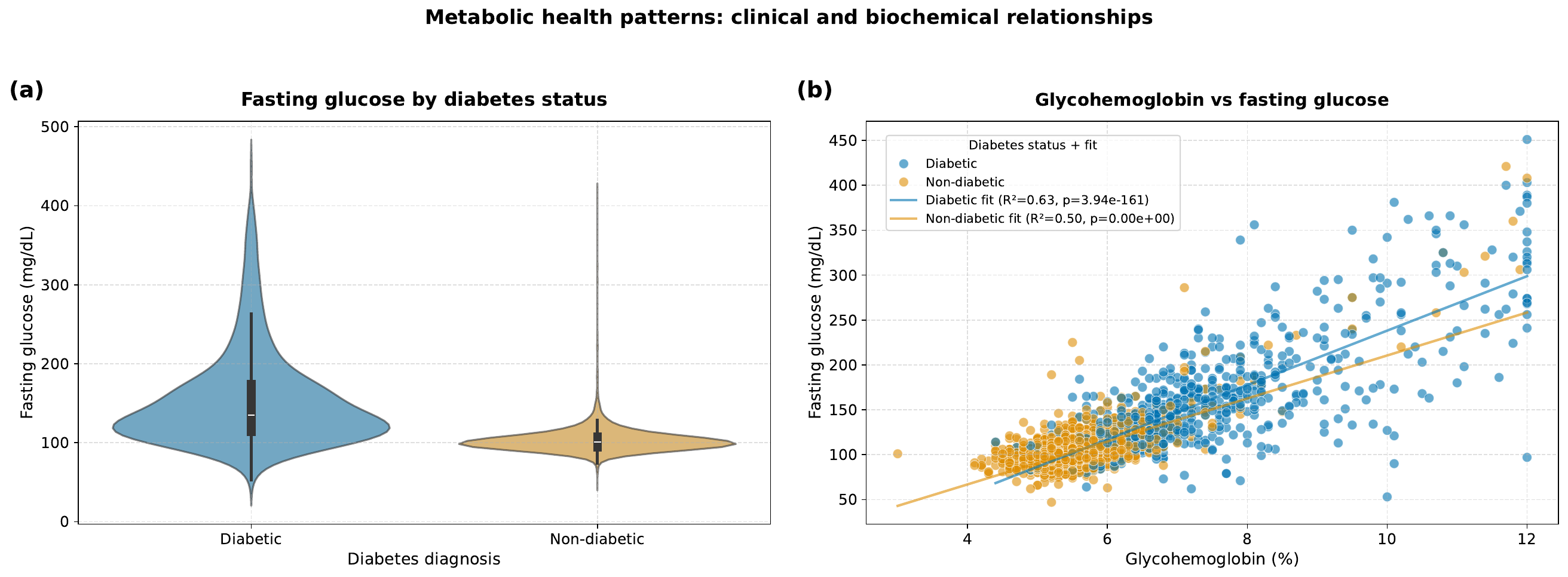}
\caption{Clinical and biochemical glucose patterns by diabetes status. (a) Violin plots showing fasting glucose distributions in diabetic and non-diabetic individuals. (b) Scatter plot of fasting glucose versus glycohemoglobin (HbA1c), coloured by diabetes status. Linear fits are shown for each group, with a stronger correlation observed in the diabetic cohort ($R^2 = 0.63$) than in the non-diabetic cohort ($R^2 = 0.50$).}
\label{fig:metabolic_relationships}
\end{figure}

As expected, fasting glucose distributions differed by diabetes status, with diabetic individuals showing both higher and more variable glucose values than non-diabetic individuals (Fig.~\ref{fig:metabolic_relationships}a). In addition, FBG and HbA1c were strongly positively associated in both groups, with a stronger correlation observed among diabetic individuals ($R^2 = 0.63$) than among non-diabetic individuals ($R^2 = 0.50$; Fig.~\ref{fig:metabolic_relationships}b). This pattern is consistent with the complementary clinical roles of these markers, as HbA1c reflects longer-term glycaemic exposure whereas FBG captures more immediate fasting glucose status. Together, these findings support the biological plausibility of the analytic sample and the suitability of FBG as the modelling target.

\begin{figure}[!t]
\centering
\includegraphics[width=0.9\columnwidth]{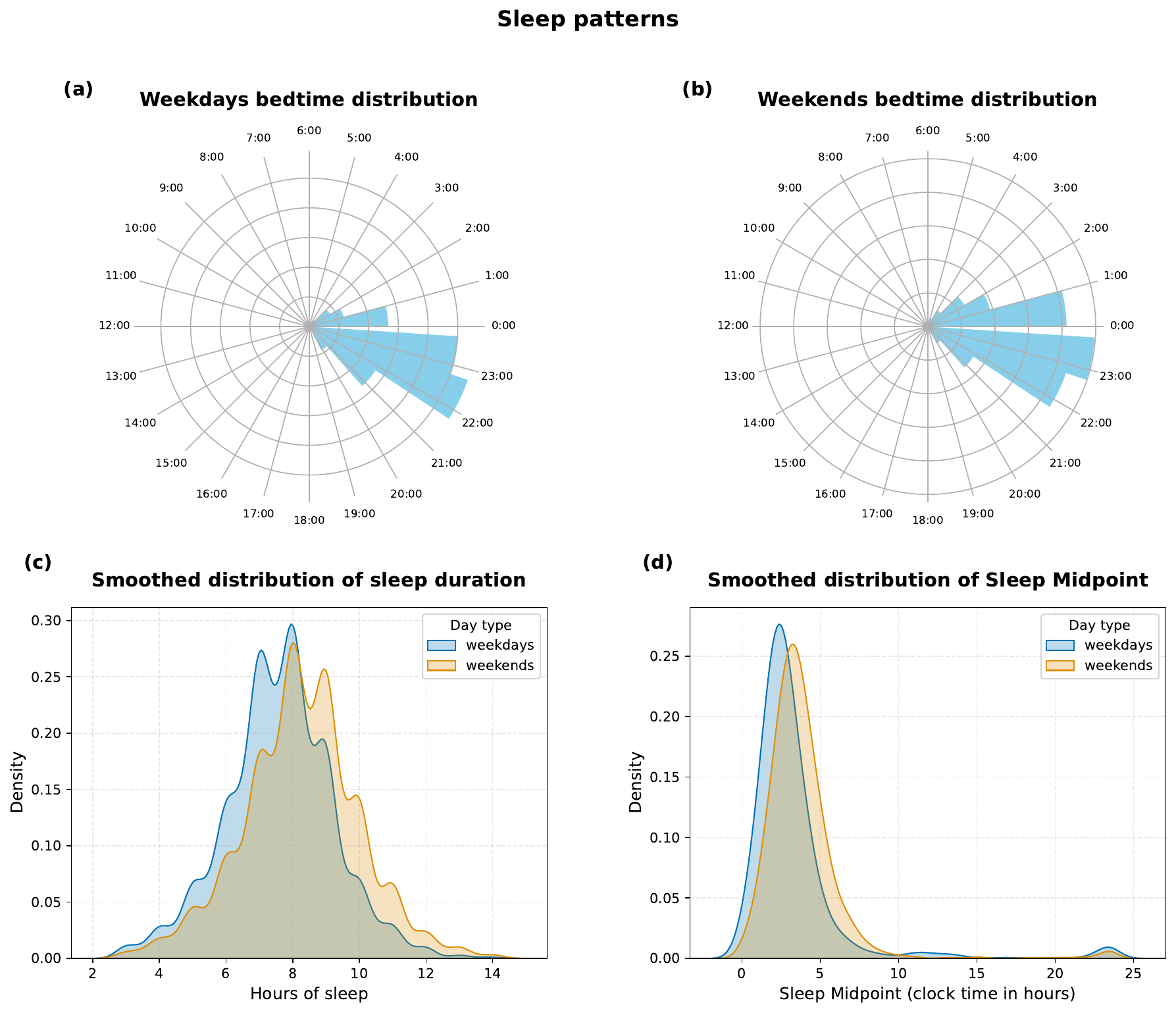}
\caption{Distribution of sleep-related variables across weekdays and weekends. (a,b) Polar histograms showing weekday and weekend bedtime distributions. (c) Smoothed density plots of sleep duration on weekdays and weekends. (d) Smoothed density plots of sleep midpoint on weekdays and weekends. Weekend distributions show later bedtimes, slightly longer sleep duration, and later sleep midpoint values, consistent with social jetlag patterns.}
\label{fig:bedtime_distribution}
\end{figure}

To assess the distribution of sleep-related circadian variables, weekday and weekend bedtime distributions, together with sleep duration and sleep midpoint, were examined (Fig.~\ref{fig:bedtime_distribution}). Polar plots showed a noticeable shift toward later bedtimes on weekends compared with weekdays, consistent with a social jetlag pattern (Fig.~\ref{fig:bedtime_distribution}a,b). Density plots further showed slightly longer sleep duration and later sleep midpoint values on weekends than on weekdays (Fig.~\ref{fig:bedtime_distribution}c,d). These patterns indicate systematic weekday--weekend differences in sleep timing within the study population and support the inclusion of circadian-related variables in subsequent modelling.

\subsection{Predictive performance of machine learning models for fasting glucose}\label{subsec:performance}

A total of 94 candidate predictors were assembled for modelling, of which 91 remained after correlation-based filtering for the full-feature models. Seven model variants were then evaluated, ranging from linear baselines to interaction-augmented XGBoost models with SHAP-guided feature selection and pruning (Table~\ref{tab:model_performance}).

\begin{table*}[t]
\caption{Performance metrics of predictive models for fasting glucose.}
\label{tab:model_performance}
\centering
\footnotesize \setlength{\tabcolsep}{4pt}
\begin{tabular}{p{0.36\textwidth}cccccc}
\toprule
\textbf{Model} & \textbf{MAE} & \textbf{RMSE} & \textbf{$R^2$ (Dev)} & \textbf{$R^2$ (Test)} & \textbf{CV $R^2$} & \textbf{Features} \\
\midrule
Model 1. Elastic Net (Full) & 0.0881 & 0.1228 & 0.6793 & 0.7441 & $0.672 \pm 0.037$ & 91 \\
Model 1b. LASSO (Full) & 0.0914 & 0.1273 & 0.6554 & 0.7251 & $0.650 \pm 0.041$ & 91 \\
Model 2. XGBoost (Full) & 0.0795 & 0.1134 & 0.7863 & 0.7817 & $0.699 \pm 0.040$ & 91 \\
Model 3. XGBoost (Selected) & 0.0813 & 0.1173 & 0.7667 & 0.7666 & $0.707 \pm 0.038$ & 12 \\
Model 4. XGBoost (+ Interactions) & 0.0805 & 0.1152 & 0.7867 & 0.7749 & $0.699 \pm 0.040$ & 95 \\
Model 5. XGBoost (+ Interactions, Selected) & 0.0805 & 0.1164 & 0.7604 & 0.7701 & $0.705 \pm 0.037$ & 12 \\
Model 5b. XGBoost (+ Interactions, Selected, Pruned) & 0.0804 & 0.1148 & 0.7532 & 0.7761 & $0.703 \pm 0.034$ & 10 \\
\bottomrule
\end{tabular}
\end{table*}

Among the linear benchmarks, Model 1 (Elastic Net, Full) outperformed Model 1b (LASSO, Full), achieving a test $R^2$ of 0.7441 and an MAE of 0.0881, compared with 0.7251 and 0.0914 for LASSO. All XGBoost-based models outperformed the linear models on the independent test set. Model 2 (XGBoost, Full) achieved the strongest overall predictive performance, with a test $R^2$ of 0.7817, an MAE of 0.0795, and an RMSE of 0.1134.

Reducing the predictor set to 12 SHAP-selected variables in Model 3 (XGBoost, Selected) led to only a modest decrease in predictive performance ($R^2 = 0.7666$, MAE $= 0.0813$), while substantially improving parsimony. Model 4 (XGBoost, + Interactions), which incorporated nine engineered interaction terms, maintained strong predictive performance ($R^2 = 0.7749$, MAE $= 0.0805$), indicating that interaction-augmented modelling preserved the predictive capacity of the full XGBoost framework. Model 5 (XGBoost, + Interactions, Selected) combined interaction-based modelling with SHAP-guided feature selection and achieved a test $R^2$ of 0.7701 with 12 predictors.

\begin{figure*}[!t]
\centering
\includegraphics[width=0.9\textwidth]{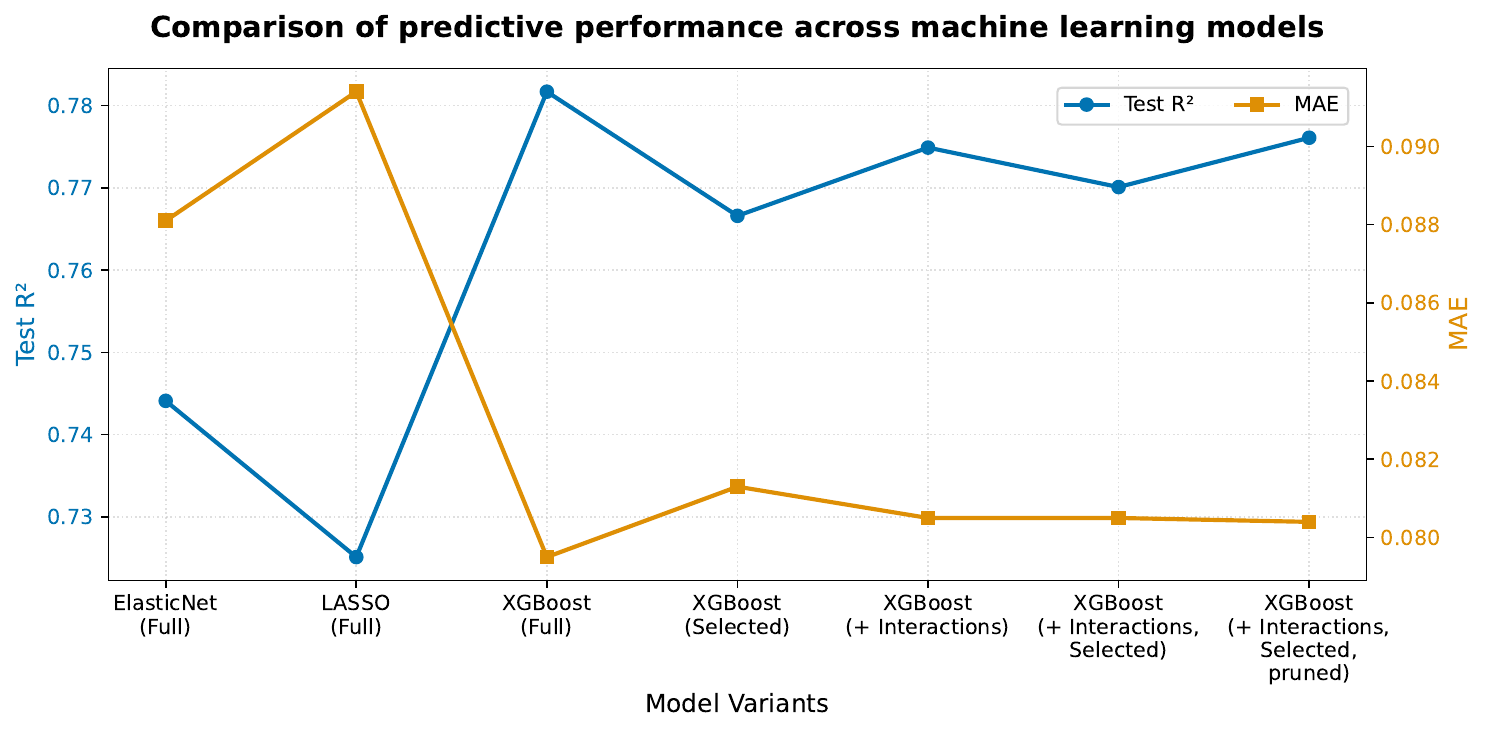}
\caption{Comparison of predictive performance across machine learning models. The blue line (left y-axis) shows test-set $R^2$, and the orange line (right y-axis) shows mean absolute error (MAE) for each model variant. Linear baselines (Elastic Net and LASSO) were outperformed by all XGBoost-based models. The full XGBoost model achieved the strongest overall test performance, whereas the final pruned interaction-augmented XGBoost model retained near-comparable accuracy using only 10 predictors, highlighting the trade-off between predictive performance and model parsimony.}
\label{fig:model_performance_evolution}
\end{figure*}

The final compact model, Model 5b (XGBoost, + Interactions, Selected, Pruned), retained only 10 predictors while achieving a test $R^2$ of 0.7761, an MAE of 0.0804, and an RMSE of 0.1148. Thus, although Model 2 yielded the highest test-set performance overall, Model 5b provided a more parsimonious solution with near-comparable predictive accuracy and improved interpretability. Cross-validated tuning results were broadly stable across model variants, and repeated random-split robustness analyses further supported the consistency of the compact interaction-augmented modelling framework. Detailed performance metrics are summarized in Table~\ref{tab:model_performance}, and Fig.~\ref{fig:model_performance_evolution} illustrates the trade-off between predictive performance and model complexity across model variants.

\subsection{Calibration of the final model}\label{subsec:calibration}

\begin{figure}[!t]
\centering
\includegraphics[width=0.9\columnwidth]{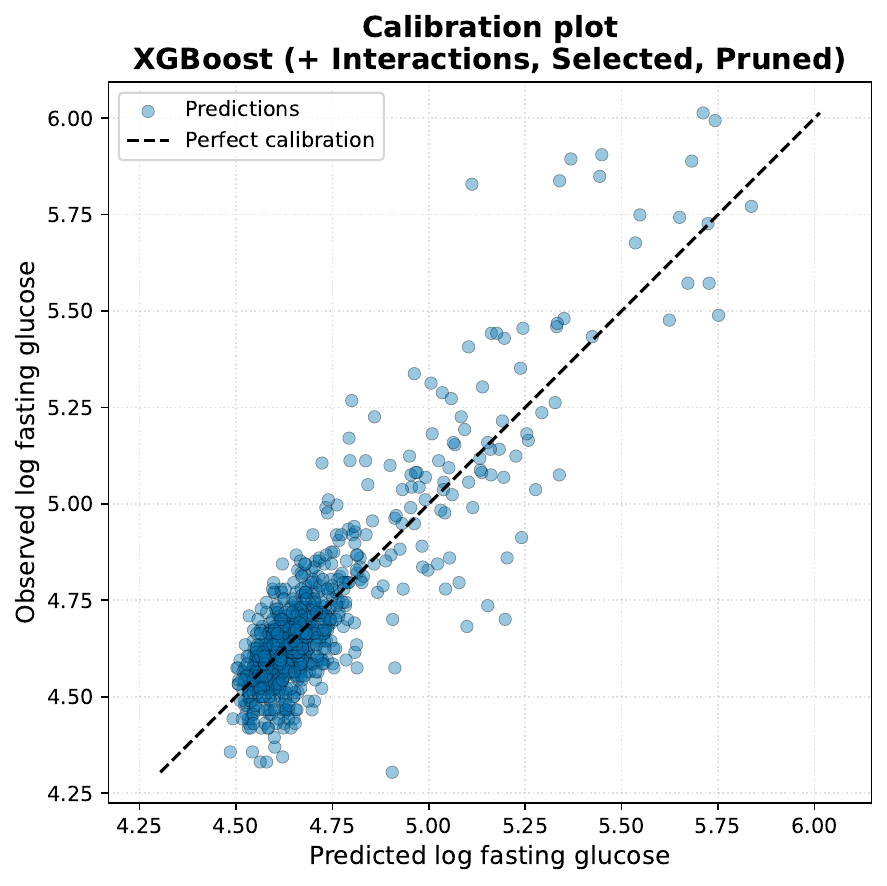}
\caption{Calibration of the final model, Model 5b (XGBoost (+ Interactions, Selected, Pruned)), on the independent test set. The scatterplot compares observed and predicted log-transformed fasting glucose values for individual participants. The dashed black line indicates perfect calibration (predicted = observed). The calibration slope (1.05) and intercept (-0.25) indicate reasonable agreement between predicted and observed values, with limited systematic bias across the prediction range.}
\label{fig:calibration}
\end{figure}

Calibration of the final model, Model 5b (XGBoost (+ Interactions, Selected, Pruned)), was assessed using both numerical calibration metrics and a calibration plot (Fig.~\ref{fig:calibration}). The calibration slope was 1.05 and the intercept was -0.25, indicating reasonable agreement between predicted and observed log-transformed fasting glucose values in the independent test set. Visual inspection of the calibration plot showed that predictions were generally well aligned with observed outcomes, with limited evidence of strong systematic bias across the prediction range.

\subsection{Predictors of fasting blood glucose}\label{subsec:predictors}

\begin{figure*}[!t]
\centering
\includegraphics[width=0.9\textwidth]{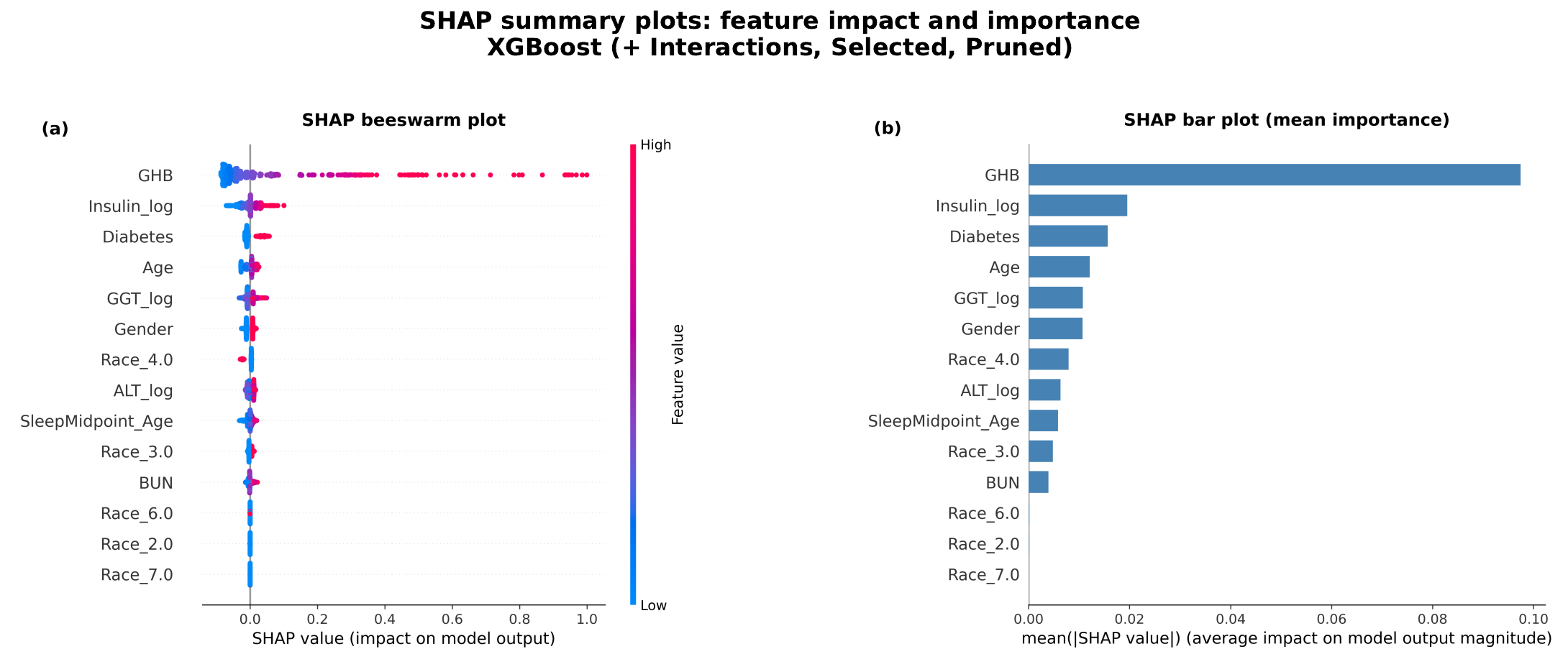}
\caption{SHAP-based feature importance for fasting glucose prediction in the final model, Model 5b (XGBoost (+ Interactions, Selected, Pruned)). (a) SHAP beeswarm plot showing the distribution and direction of feature effects across individual predictions, with points coloured by feature value. (b) Mean absolute SHAP values summarizing each feature’s average contribution to model output. GHB was the dominant predictor, followed by insulin, diabetes diagnosis, GGT, age, race, gender, and the \textit{Sleep Midpoint} $\times$ \textit{Age} interaction.}
\label{fig:shap}
\end{figure*}

SHAP analysis was used to quantify feature importance and interpret model predictions at both the global and local levels for the final model, Model 5b (XGBoost (+ Interactions, Selected, Pruned)) (Fig.~\ref{fig:shap}). The SHAP beeswarm plot (Fig.~\ref{fig:shap}a) summarizes the distribution and direction of feature effects across all predictions, whereas the SHAP bar plot (Fig.~\ref{fig:shap}b) shows mean absolute SHAP values as a measure of average feature importance. In the final model, glycohemoglobin (HbA1c; labelled as GHB in model figures) emerged as the dominant predictor, with the largest mean absolute SHAP value (0.0975). This is consistent with its established biological relationship to longer-term glycaemic exposure: glycohemoglobin provides a blood glucose level reflection for the past 120 days, which should be very closely related to daily fasting blood glucose levels. Other highly ranked predictors included iinsulin, diabetes diagnosis, age, race, and gender, all of which are known to be associated with glucose metabolism and metabolic dysfunction. Another significant contributor, a liver enzyme GGT, was linked to fasting blood glucose levels and was even proposed as an early detection marker for metabolic dysfunction~\cite{tengAssociationSerumGamma2023}. The prominence of these variables supports the biological plausibility of the final model.

Interestingly, among the engineered interaction terms, only \textit{Sleep Midpoint} $\times$ \textit{Age} ranked among the top predictors(Fig.~\ref{fig:shap}). Although its contribution was modest relative to dominant glycaemic and metabolic markers, its repeated retention in the full-model interaction pipeline suggests a small but stable age-modulated circadian signal within the predictive framework. This result is notable given that prior machine-learning studies of glycaemic prediction have rarely incorporated circadian timing features together with explicit interaction terms.

To further assess the stability of feature ranking, permutation importance was computed on the independent test set (Appendix Table~\ref{tab:perm_importance}). The resulting ordering closely mirrored the SHAP results, with GHB remaining the dominant predictor, followed by insulin, diabetes diagnosis, GGT, age, race, and gender. The \textit{Sleep Midpoint} $\times$ \textit{Age} interaction remained among the retained predictors, although with a comparatively small mean permutation importance (0.0003), indicating that its contribution was substantially weaker than that of the dominant metabolic features.

\begin{figure*}[!t]
\centering
\includegraphics[width=0.9\textwidth]{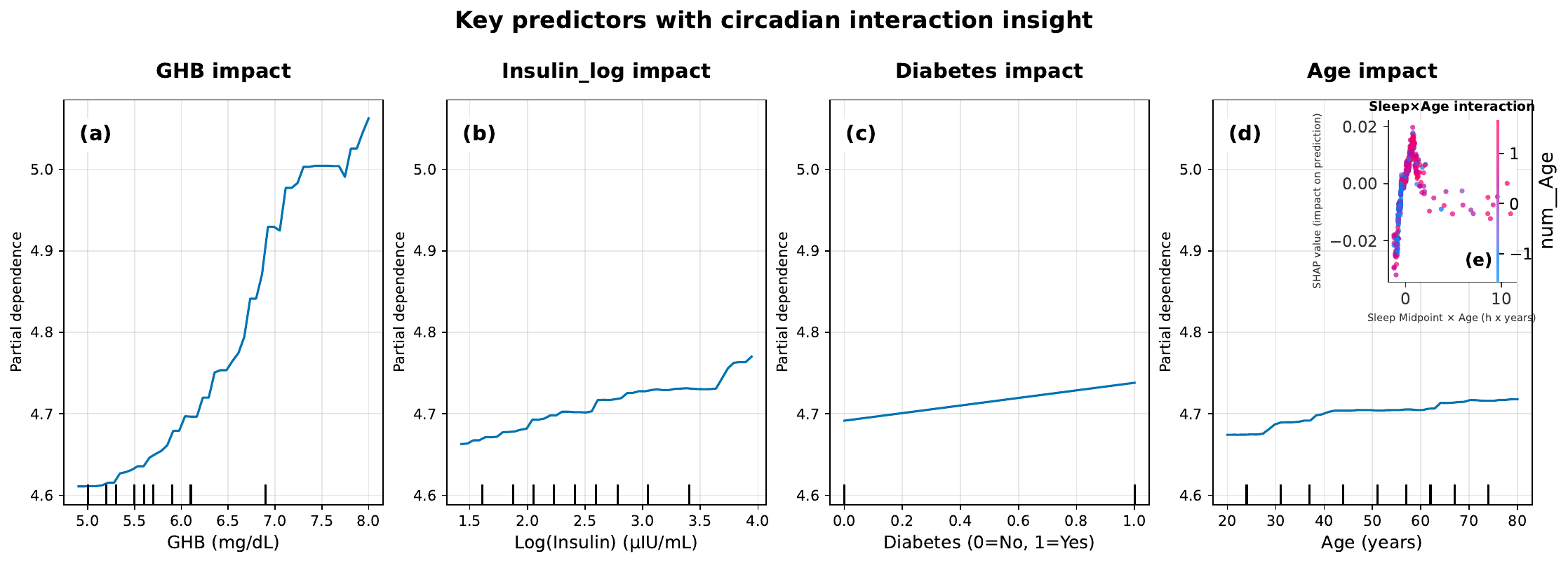}
\caption{Partial dependence plots for key predictors in the final model, Model 5b (XGBoost (+ Interactions, Selected, Pruned)). Partial dependence plots show the marginal association of (a) GHB, (b) insulin, (c) diabetes diagnosis, and (d) age with predicted fasting glucose. The inset (e) in panel (d) shows SHAP values for the \textit{Sleep Midpoint} $\times$ \textit{Age} interaction, highlighting its nonlinear contribution to model predictions.}\label{fig:pdp}
\end{figure*}

Partial dependence plots were used to visualize the marginal effects of several key predictors on model output (Fig.~\ref{fig:pdp}). Higher GHB values were associated with higher predicted fasting glucose (Fig.~\ref{fig:pdp}a), and predicted fasting glucose also increased gradually with increasing insulin levels (Fig.~\ref{fig:pdp}b). Participants with diabetes diagnosis had higher predicted fasting glucose values than non-diabetic participants (Fig.~\ref{fig:pdp}c), and age showed a modest positive association with predicted glucose values (Fig.~\ref{fig:pdp}d). These trends are consistent with established physiological expectations.

\begin{figure}[!t]
\centering
\includegraphics[width=0.9\columnwidth]{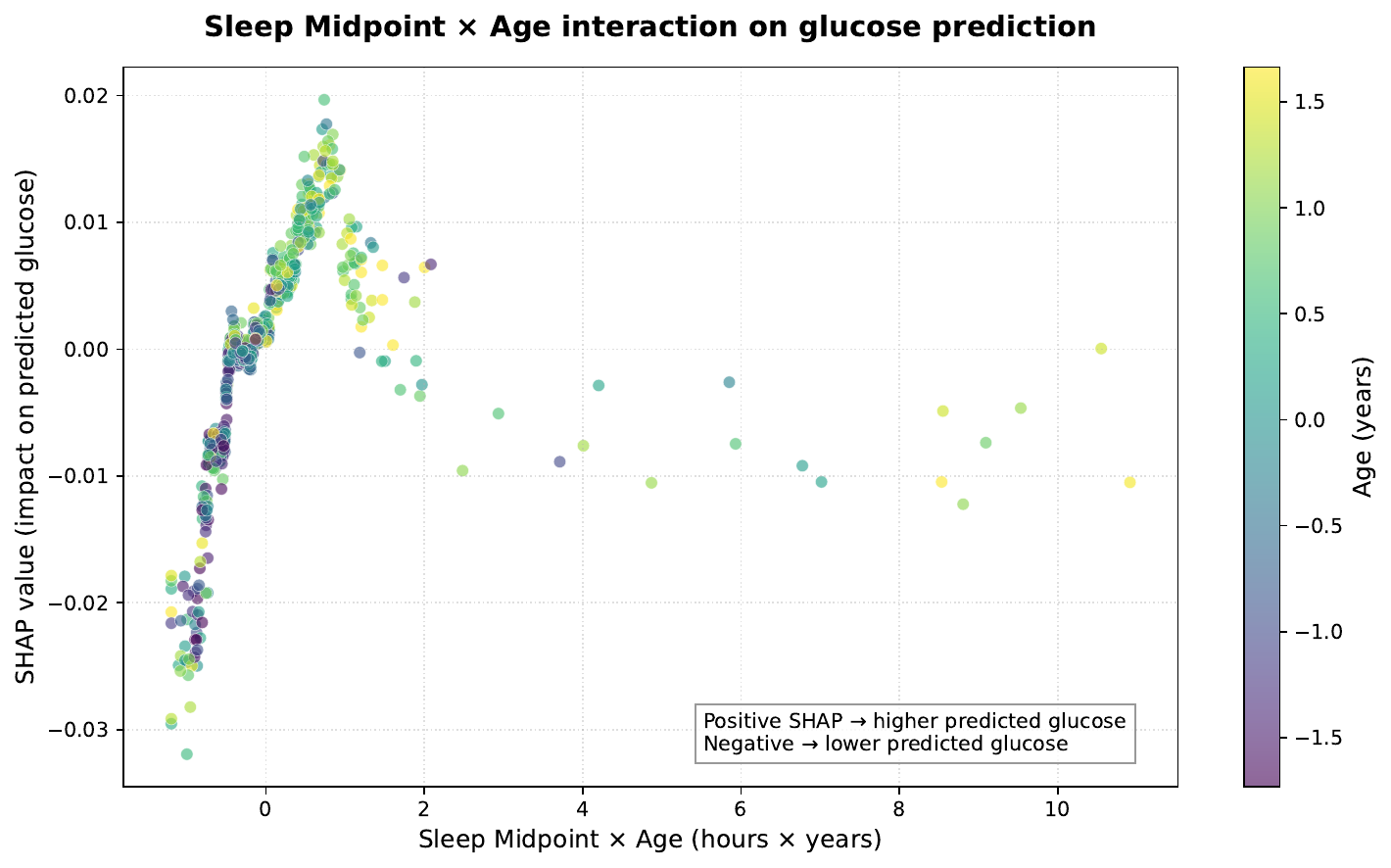}
\caption{SHAP dependence plot for the \textit{Sleep Midpoint} $\times$ \textit{Age} interaction in the final model, Model 5b (XGBoost (+ Interactions, Selected, Pruned)). Each point represents an individual prediction and is colour-coded by age. Positive SHAP values indicate higher predicted fasting glucose, whereas negative SHAP values indicate lower predicted fasting glucose. The pattern suggests a nonlinear interaction between sleep timing and age in relation to model-predicted fasting glucose.}
\label{fig:sleep_age_interaction}
\end{figure}

The \textit{Sleep Midpoint} $\times$ \textit{Age} interaction showed a more complex pattern than the main-effect predictors (Fig.~\ref{fig:pdp}e; Fig.~\ref{fig:sleep_age_interaction}). SHAP values for this interaction varied nonlinearly across the observed range. In general, the interaction appeared to contribute more strongly among older individuals at moderate values of the term, whereas very large values were sparse and more variable. This pattern suggests that the relationship between sleep timing and glucose regulation may be modified by age, although the effect size was modest relative to dominant glycaemic predictors. The observed shape is consistent with emerging evidence that age-related circadian disruption may influence glucose regulation \cite{zhangCircadianRhythmGlucose2025, xuRestactivityCircadianRhythm2022}.

\subsection{Repeated random-split robustness of interaction-augmented models}\label{subsec:robustness}

To assess the stability of the interaction-augmented modelling pipeline, repeated random-split robustness analyses were conducted across 10 random seeds (0, 1, 2, 3, 4, 5, 10, 20, 42, and 100) for Model 4, Model 5, and Model 5b. Across seeds, predictive performance remained broadly stable for all three models (Appendix Table~\ref{tab:robustness_performance}). Model 4 achieved a mean test $R^2$ of 0.742 $\pm$ 0.036 with a mean MAE of 0.0826 $\pm$ 0.0022, whereas Model 5 achieved a mean test $R^2$ of 0.744 $\pm$ 0.033 and a mean MAE of 0.0824 $\pm$ 0.0023. The final compact model, Model 5b, showed similar stability, with a mean test $R^2$ of 0.746 $\pm$ 0.033, a mean MAE of 0.0823 $\pm$ 0.0023, and a mean RMSE of 0.1207 $\pm$ 0.0068. Mean tuned cross-validated $R^2$ values were also similar across the three models, indicating that the compact interaction-augmented models preserved predictive performance while reducing model complexity.

Feature-selection stability was examined across the repeated random splits for the SHAP-guided reduced interaction models. Several predictors were selected consistently in all runs, including GHB, insulin, diabetes diagnosis, age, race, GGT, and gender. Notably, the engineered interaction term \textit{Sleep Midpoint} $\times$ \textit{Age} was retained in 100\% of repeated runs for both Model 5 and Model 5b, supporting its stability within the full-model interaction pipeline (Appendix Table~\ref{tab:robustness_feature_frequency_importance}). However, its mean permutation importance in Model 5b remained small (mean = 0.0020, SD = 0.0012), indicating that this stable interaction contributed modestly relative to dominant glycaemic and metabolic predictors. Together, these repeated random-split and permutation-importance analyses provide an empirical measure of variability around the final-model importance ranking and support a cautious interpretation of the SHAP-guided feature hierarchy.

A repeated random-split ablation analysis comparing Model 5b with a corresponding refitted model excluding \textit{Sleep Midpoint} $\times$ \textit{Age} (Appendix Table~\ref{tab:ablation_sleepmid}) showed negligible differences in predictive performance across 10 seeds (mean test $R^2$: 0.7459 vs.\ 0.7456; mean MAE: 0.08230 vs.\ 0.08235), further supporting the interpretation that this interaction was stable but modest rather than a major driver of predictive gain.

\subsection{Restricted sensitivity analyses excluding major glycaemic predictors}\label{subsec:restricted_sensitivity}

To address the dependence of model performance on dominant glycaemic predictors, restricted sensitivity analyses were conducted after excluding glycohemoglobin (GHB/HbA1c), insulin, diabetes status, and related interaction terms from the candidate feature set. As expected, predictive performance decreased substantially across the restricted interaction-augmented models (Appendix Table~\ref{tab:restricted_performance}). Mean test-set $R^2$ values across 10 random seeds were 0.354 for restricted Model 4, 0.338 for restricted Model 5, and 0.340 for restricted Model 5b, with corresponding MAE values of approximately 0.127--0.128. In this restricted setting, the repeatedly selected predictors were primarily age, renal and liver-related markers, lipids, and hormonal variables, and the \textit{Sleep Midpoint} $\times$ \textit{Age} interaction was not retained in any repeated run. These results indicate that the circadian-age interaction observed in the full-model framework was context-dependent rather than dominant in the absence of major glycaemic predictors.

\section{Discussion}

This study developed an interpretable machine learning for predicting fasting glucose using a broad set of metabolic, hormonal, circadian, lifestyle, nutritional, and demographic variables derived from NHANES 2017--2020 pre-pandemic data. The large, heterogeneous analytic sample supported robust model development and evaluation, and included both diabetic and non-diabetic individuals with biologically plausible distributions of fasting glucose and glycohemoglobin (HbA1c; labelled as GHB in model figures)  (Fig.~\ref{fig:metabolic_relationships}). Compared with many previous fasting glucose prediction studies, the present framework incorporated a wider range of candidate predictors, including sleep- and circadian-related variables (Table~\ref{tab:variables}), and explicitly evaluated interaction-augmented models. The final compact model, Model 5b (XGBoost (+ Interactions, Selected, Pruned)), achieved strong predictive performance while remaining interpretable, with a test MAE of 0.0804, RMSE of 0.1148, and test $R^2$ of 0.7761. After back-transformation to the original fasting-glucose scale, this corresponded to a mean absolute error of approximately 10.3\,mg/dL and a root mean squared error of approximately 18.8\,mg/dL on the independent test set.

Among all predictors, GHB consistently emerged as the dominant feature, followed by insulin, diabetes diagnosis, GGT, age, race, and gender. These variables have well-established links to glucose metabolism and metabolic dysfunction, which supports the biological plausibility of the final model. As expected, the strongest predictors were conventional glycaemic and metabolic markers that are closely related to fasting glucose physiology. Within this broader predictive context, SHAP-based analyses indicated that circadian-related information also entered the final compact model, particularly through the \textit{Sleep Midpoint} $\times$ \textit{Age} interaction (Fig.~\ref{fig:shap}). Although its magnitude was modest relative to dominant metabolic predictors, this interaction was consistently retained in the compact interaction-augmented full-model pipeline and showed a nonlinear SHAP pattern suggestive of age-modified sleep-timing effects on model-predicted fasting glucose.

The repeated random-split robustness analyses provide important context for interpreting this interaction. Across 10 random seeds, the compact interaction-augmented models showed stable predictive performance, and \textit{Sleep Midpoint} $\times$ \textit{Age} was retained in 100\% of repeated runs for both Model 5 and Model 5b. This supports the reproducibility of the interaction within the full-model pipeline. At the same time, its permutation importance remained small relative to dominant glycaemic predictors, indicating that it should be interpreted as a modest but stable contributor rather than a major standalone driver of model performance.

The restricted sensitivity analyses further showed that when major glycaemic predictors and related interaction terms were removed, predictive performance declined substantially and the \textit{Sleep Midpoint} $\times$ \textit{Age} interaction was no longer retained, indicating that its contribution was context-dependent rather than dominant in the absence of strong glycaemic information.

Previous studies have linked circadian misalignment, sleep timing, impaired glucose tolerance, and diabetes risk~\cite{morrisEffectsInternalCircadian2016, morrisEndogenousCircadianSystem2015, wollerCircadianMisalignmentMetabolic2021, shenTrajectoriesSleepDuration2025, javeedCircadianEtiologyType2018}. More recent machine learning studies have begun to incorporate chronobiological features or wearable-derived circadian metrics into glycaemic prediction~\cite{phillipsUncoveringPersonalizedGlucose2023, burksChronobiologicallyinformedFeaturesCGM2025}. However, these approaches have typically treated circadian variables as independent predictors rather than explicitly modelling their interaction with demographic factors such as age~\cite{sadriaAgingAffectsCircadian2021, rothCircadianmediatedRegulationCardiometabolic2023}. In this context, the present study extends the literature by showing that an age-modulated sleep-timing feature can be stably retained in an interpretable multivariable prediction framework, even though its effect is modest relative to dominant clinical markers.

Several limitations should be considered. First, the cross-sectional nature of NHANES prevents causal inference and limits assessment of temporal relationships between circadian timing and glucose regulation. Second, sleep midpoint is only a proxy for circadian alignment, and more direct physiological markers, such as melatonin or cortisol, would provide a more precise characterization of circadian phase~\cite{zhangCircadianRhythmGlucose2025}; however, such biomarkers were not available in NHANES dataset. Third, sleep-related variables were derived from questionnaire-based measures and therefore may not capture circadian phase or sleep behaviour with high precision. Fourth, external validation on an independent NHANES cycle or another cohort was not performed, which limits assessment of generalizability across populations and time periods. Fifth, NHANES survey weights were not incorporated, so the findings should be interpreted as internally validated predictive results within the assembled analytic sample rather than nationally weighted estimates of association or prediction performance. In addition, the strongest predictors in the full models were glycaemic markers that are biologically close to the outcome, which is valuable for predictive optimisation but limits the extent to which the model can be interpreted as identifying independent circadian determinants of fasting glucose. Finally, although the study included a broad and diverse U.S. sample, the results may not generalize fully to non-NHANES populations or to individuals with more severe metabolic disease profiles.

Future work should evaluate similar models in longitudinal datasets, incorporate more direct circadian biomarkers where available, and test generalizability in external cohorts. It would also be valuable to investigate whether circadian features provide greater utility in settings with fewer laboratory predictors, in non-laboratory-based models, or in prospective risk-prediction frameworks. Overall, the present study shows that interpretable machine learning can recover both dominant metabolic predictors and weaker circadian-related signals within a unified framework. In particular, the \textit{Sleep Midpoint} $\times$ \textit{Age} interaction appears to represent a stable but modest circadian-age signal that warrants further investigation in metabolic health research. 

\subsection*{Code availability}
The full Python codebase, including all data processing, feature engineering, model pipelines, and analysis scripts, is openly available on GitHub: 
\url{https://github.com/Vitashka1995/circadian-glucose-prediction.git}

\appendix

\section{Sample construction and missing-data summary}\label{secA1}
This appendix summarizes the main sample-construction steps and the pattern of missing data in the final analytic dataset used for modelling. As described in the main text, missing-data handling followed a staged strategy. Participants with missing fasting glucose, missing critical sleep variables required for circadian-feature derivation, or more than 50\% missingness across the merged biomarker dataset were excluded before modelling. The remaining missing predictor values were then handled within the machine-learning pipeline using imputation. Table~\ref{tab:sample_flow} summarizes the main filtering steps.

After these exclusions, missingness was zero for all retained circadian variables and for the lifestyle and medical-history variables derived from the sleep and activity questionnaires. Among control variables, family income-to-poverty ratio showed 13.02\% missingness. Among hormonal variables, missingness ranged from 11.95\% to 15.09\%. Within the metabolic domain, missingness was generally low to moderate, ranging from 0.27\% for glycohemoglobin (HbA1c; GHB) to 8.89\% for systolic and diastolic blood pressure. Nutritional variables derived from the dietary recall data showed 6.72\% missingness, whereas the dietary indicator variables and the fasting-glucose outcome had no missing values in the final analytic dataset.

\begin{table}[!t]
\caption{Construction of the final analytic sample.}
\label{tab:sample_flow}
\centering
\begin{tabular}{p{0.68\columnwidth}r}
\toprule
\textbf{Sample construction step} & \textbf{N} \\
\midrule
Participants in the biomarker merge base with fasting glucose and insulin data available for merging & 5,090 \\
Excluded: rows with more than 50\% missingness across the merged biomarker dataset & 145 \\
Remaining biomarker dataset after row-level missingness filtering & 4,945 \\
\midrule
Participants in the sleep questionnaire dataset before filtering & 10,195 \\
Excluded: missing critical sleep variables required to derive circadian measures & 222 \\
Remaining sleep dataset after critical-variable filtering & 9,973 \\
\midrule
Final merged analytic sample used for modelling & 4,016 \\
\bottomrule
\end{tabular}
\end{table}

\section{Code for sleep and circadian feature engineering}\label{secA2}

\begin{figure}[!t]
\centering
\includegraphics[width=\columnwidth]{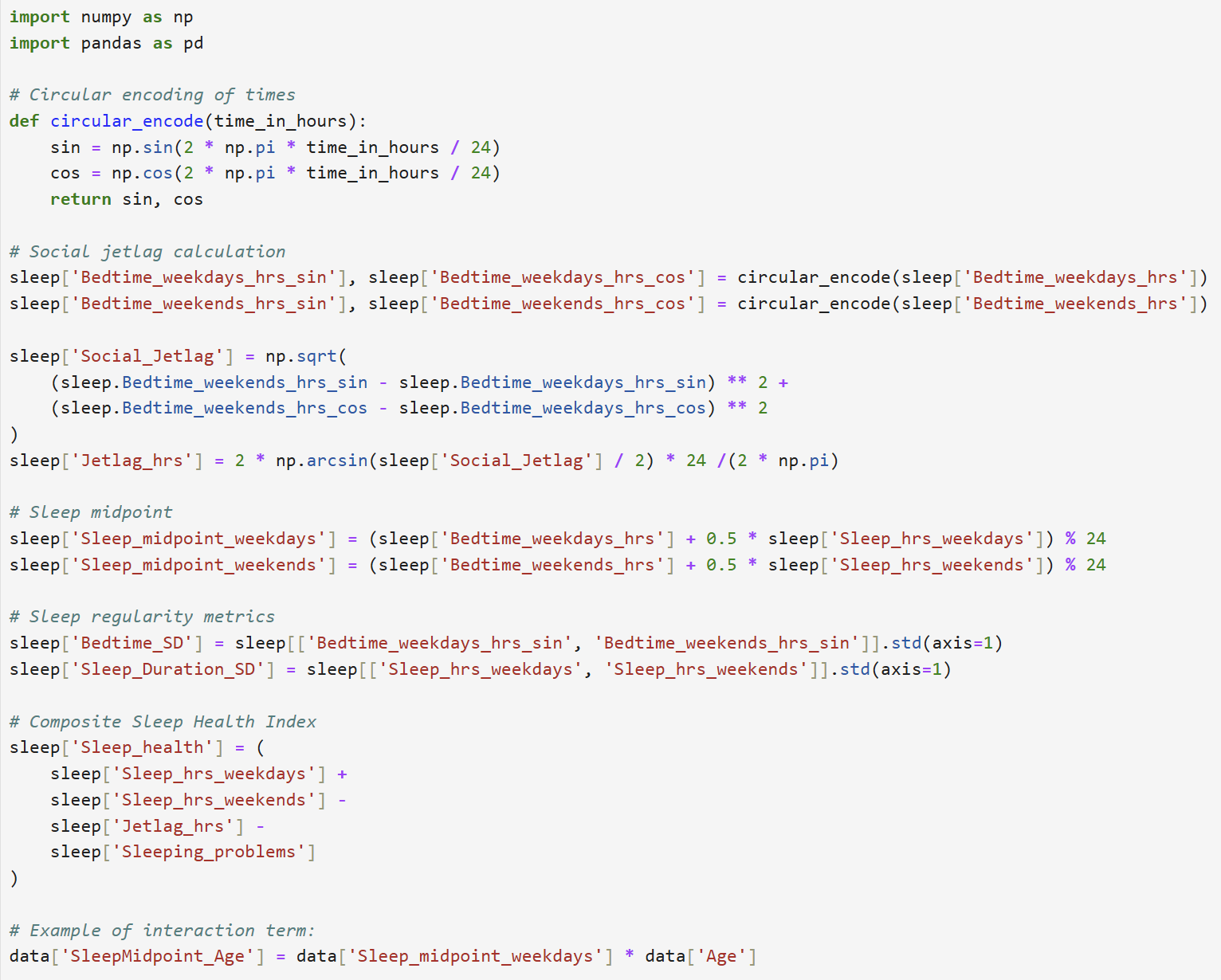}
\caption{Python code used to derive engineered sleep and circadian-related variables, including circular encoding, social jetlag, sleep midpoint, sleep regularity metrics, the composite sleep health index, and an example interaction term.}
\label{fig:supp_code_sleep}
\end{figure}

Appendix Fig.~\ref{fig:supp_code_sleep} shows the Python code used to derive the engineered sleep and circadian-related variables described in the main Methods. These include circular encoding of time variables, social jetlag, sleep midpoint, sleep regularity metrics, the composite sleep health index, and an example interaction term used in the predictive models.

\section{Hierarchical correlation structure of metabolic and circadian-related features}\label{secA3}

\begin{figure*}[t]
\centering
\includegraphics[width=0.9\textwidth]{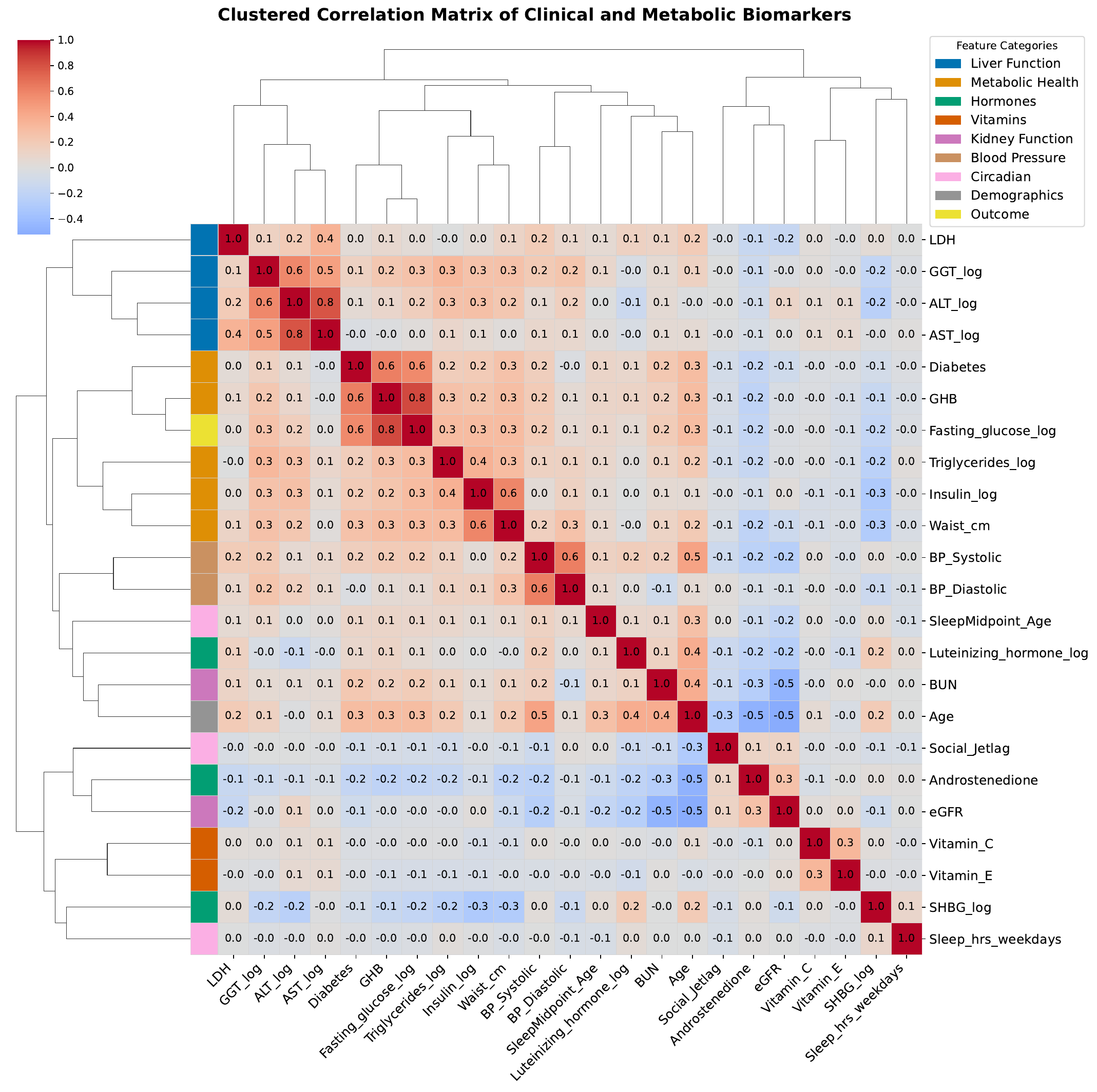}
\caption{Clustered correlation matrix of clinical, metabolic, hormonal, circadian, and demographic features. Colours range from blue (negative correlation) to red (positive correlation), centred at zero. Hierarchical clustering of the Pearson correlation matrix reveals groups of related variables across physiological and behavioural domains. Row and column sidebars are colour-coded by feature category, illustrating the multidimensional structure of the dataset and the relationships among metabolic and circadian-related variables.}\label{fig:cluster_heatmap}
\end{figure*}

Appendix Fig.~\ref{fig:cluster_heatmap} shows a clustered heatmap of Pearson correlations among standardised features, revealing groupings across physiological and behavioural domains. Sidebars annotate variables by feature category, including liver function, metabolic health, hormones, circadian variables, demographics, kidney function, blood pressure, vitamins, and the outcome. As expected, a cluster of metabolically related variables emerged, including waist circumference, triglycerides, glycohemoglobin (HbA1c), insulin, and fasting glucose, consistent with known insulin resistance-related patterns. In addition, the \textit{Sleep Midpoint} $\times$ \textit{Age} interaction clustered near age and several metabolic features, supporting its consideration as a candidate circadian-age term in the predictive modelling framework. Overall, the correlation structure supports the inclusion of both conventional metabolic markers and circadian-related variables in the model-development pipeline.

\section{Hyperparameter tuning procedures and optimal model settings}\label{secA4}

Tables~\ref{tab:hyperparameter_space} and~\ref{tab:optimal_hyperparameters} present the hyperparameter search spaces and corresponding optimal settings for all predictive models evaluated in this study. Elastic Net and LASSO hyperparameters were optimized using 20 randomized-search iterations, whereas XGBoost models were optimized using 50 iterations. In all cases, hyperparameter tuning was performed within the model-development subset using five-fold cross-validated randomized search, with $R^2$ used as the optimization metric. For XGBoost-based reduced-feature models, the reported optimal settings correspond to the final tuning run performed after SHAP-guided feature selection within the model-development subset.

\begin{table}[t]
\caption{Hyperparameter search spaces used for model tuning.}
\label{tab:hyperparameter_space}
\centering
\begin{tabular}{p{0.28\columnwidth} p{0.62\columnwidth}}
\toprule
\textbf{Model} & \textbf{Hyperparameter search space} \\
\midrule
Model 1: Elastic Net (Full) &
$\alpha \sim \mathrm{Uniform}(0.001, 1)$; \newline
$l1\_ratio \sim \mathrm{Uniform}(0, 1)$ \\
\midrule
Model 1b: LASSO (Full) &
$\alpha \sim \mathrm{Uniform}(0.0005, 1)$ \\
\midrule
Models 2--5b: XGBoost &
$n\_estimators \in \{1000, 1200, 1500\}$; \newline
$learning\_rate \in \{0.01, 0.015, 0.02\}$; \newline
$max\_depth \in \{2, 3\}$; \newline
$min\_child\_weight \in \{10, 12, 15\}$; \newline
$subsample \in \{0.7, 0.8\}$; \newline
$colsample\_bytree \in \{0.7, 0.8\}$; \newline
$reg\_alpha \in \{0.1, 0.2\}$; \newline
$reg\_lambda \in \{10, 15, 20\}$; \newline
$gamma \in \{0.05, 0.1\}$ \\
\bottomrule
\end{tabular}
\end{table}

\begin{table}[t]
\caption{Optimal hyperparameter settings identified during model-development tuning.}
\label{tab:optimal_hyperparameters}
\centering
\begin{tabular}{p{0.36\columnwidth} p{0.54\columnwidth}}
\toprule
\textbf{Model} & \textbf{Optimal settings} \\
\midrule
Model 1: Elastic Net (Full) &
$\alpha = 0.0720$; $l1\_ratio = 0.0871$ \\
\midrule
Model 1b: LASSO (Full) &
$\alpha = 0.0207$ \\
\midrule
Model 2: XGBoost (Full) &
$subsample = 0.7$; $reg\_lambda = 10$; $reg\_alpha = 0.1$; $n\_estimators = 1000$; $min\_child\_weight = 12$; $max\_depth = 2$; $learning\_rate = 0.02$; $gamma = 0.05$; $colsample\_bytree = 0.8$ \\
\midrule
Model 3: XGBoost (Selected) &
$subsample = 0.7$; $reg\_lambda = 15$; $reg\_alpha = 0.1$; $n\_estimators = 1200$; $min\_child\_weight = 10$; $max\_depth = 2$; $learning\_rate = 0.02$; $gamma = 0.05$; $colsample\_bytree = 0.7$ \\
\midrule
Model 4: XGBoost (+ Interactions) &
$subsample = 0.8$; $reg\_lambda = 10$; $reg\_alpha = 0.2$; $n\_estimators = 1200$; $min\_child\_weight = 15$; $max\_depth = 3$; $learning\_rate = 0.01$; $gamma = 0.1$; $colsample\_bytree = 0.8$ \\
\midrule
Model 5: XGBoost (+ Interactions, Selected) &
$subsample = 0.7$; $reg\_lambda = 15$; $reg\_alpha = 0.2$; $n\_estimators = 1000$; $min\_child\_weight = 12$; $max\_depth = 2$; $learning\_rate = 0.02$; $gamma = 0.05$; $colsample\_bytree = 0.8$ \\
\midrule
Model 5b: XGBoost (+ Interactions, Selected, Pruned) &
$subsample = 0.7$; $reg\_lambda = 15$; $reg\_alpha = 0.2$; $n\_estimators = 1000$; $min\_child\_weight = 12$; $max\_depth = 2$; $learning\_rate = 0.02$; $gamma = 0.05$; $colsample\_bytree = 0.8$ \\
\bottomrule
\end{tabular}
\end{table}

\begin{table}[htbp]
\caption{Repeated random-split robustness summary across 10 random seeds for interaction-augmented XGBoost models.}
\label{tab:robustness_performance}
\centering

\begin{tabular*}{\textwidth}{
@{\extracolsep{\fill}}
>{\raggedright\arraybackslash}p{0.28\textwidth}
cccc
@{}
}
\toprule
\textbf{Model} & \textbf{Test $R^2$} & \textbf{MAE} & \textbf{RMSE} & \textbf{Tuned CV $R^2$} \\
\midrule

Model 4: XGBoost (+ Interactions) 
& $0.742 \pm 0.036$ & $0.0826 \pm 0.0022$ & $0.1215 \pm 0.0071$ & $0.7079 \pm 0.0105$ \\

Model 5: XGBoost (+ Interactions, Selected)
& $0.744 \pm 0.033$ & $0.0824 \pm 0.0023$ & $0.1212 \pm 0.0065$ & $0.7136 \pm 0.0100$ \\

Model 5b: XGBoost (+ Interactions, Selected, Pruned)
& $0.746 \pm 0.033$ & $0.0823 \pm 0.0023$ & $0.1207 \pm 0.0068$ & $0.7138 \pm 0.0096$ \\

\bottomrule
\end{tabular*}

\vspace{1mm}

\parbox{\textwidth}{%
\footnotesize
\emph{Note:} Values are reported as mean $\pm$ standard deviation across seeds.
}

\end{table}

\section{Permutation feature importance of the final model}\label{secA5}

Table~\ref{tab:perm_importance} reports permutation feature importance scores for the final model, Model 5b (XGBoost (+ Interactions, Selected, Pruned)). Mean importance and standard deviation were estimated using 10 repeated shuffles per feature on the independent test set, providing an additional measure of variability in the final-model feature ranking.

\begin{table}[t]
\caption{Permutation feature importance scores for the final model, Model 5b (XGBoost (+ Interactions, Selected, Pruned)).}
\label{tab:perm_importance}
\centering
\begin{tabular}{lcc}
\toprule
\textbf{Feature} & \textbf{Importance (mean)} & \textbf{Importance (SD)} \\
\midrule
GHB                          & 1.0973 & 0.0252 \\
Insulin (log)                & 0.0186 & 0.0018 \\
Diabetes                     & 0.0155 & 0.0021 \\
GGT (log)                    & 0.0121 & 0.0035 \\
Age                          & 0.0117 & 0.0025 \\
Race                         & 0.0097 & 0.0022 \\
Gender                       & 0.0036 & 0.0021 \\
BUN                          & 0.0030 & 0.0012 \\
ALT (log)                    & 0.0009 & 0.0006 \\
Sleep Midpoint $\times$ Age  & 0.0003 & 0.0012 \\
\bottomrule
\end{tabular}

\vspace{1mm}
\footnotesize \emph{Note:} Mean and standard deviation were computed using 10 repeated shuffles per feature on the independent test set and are shown for the 10 retained predictors.
\end{table}

\section{Repeated random-split robustness analyses of interaction-augmented models}\label{secA6}

To evaluate the stability of the interaction-augmented modelling pipeline beyond a single train-test split, repeated random-split robustness analyses were conducted across 10 random seeds (0, 1, 2, 3, 4, 5, 10, 20, 42, and 100) for Model 4, Model 5, and Model 5b. For each seed, the dataset was re-split into model-development and independent test sets using the same age-stratified procedure as in the main analysis, followed by hyperparameter tuning, SHAP-guided feature selection, pruning, and final evaluation. Table~\ref{tab:robustness_performance} summarizes the distribution of predictive performance across seeds, whereas Table~\ref{tab:robustness_feature_frequency_importance} reports feature-selection frequency and across-seed permutation importance summaries for the final compact model pipeline (Model 5b). Table~\ref{tab:ablation_sleepmid} reports a repeated random-split targeted ablation analysis in which \textit{Sleep Midpoint} $\times$ \textit{Age} was removed from the final compact model to quantify its incremental contribution to predictive performance. These analyses were designed to assess model-performance stability, reproducibility of the selected predictor set, and the robustness of the key circadian-age interaction across repeated random splits.

\begin{table}[t]
\caption{Feature-selection stability and across-seed permutation-importance summary for the final compact interaction-augmented model pipeline (Model 5b).}
\label{tab:robustness_feature_frequency_importance}
\centering
\footnotesize
\setlength{\tabcolsep}{3pt}
\begin{tabular}{p{0.34\columnwidth}ccc}
\toprule
\textbf{Feature} & \textbf{Count} & \textbf{Frequency} & \makecell[c]{\textbf{Importance}\\\textbf{mean $\pm$ SD}} \\
\midrule
GHB & 10 & 1.0 & $1.0585 \pm 0.0635$ \\
Insulin (log) & 10 & 1.0 & $0.0278 \pm 0.0048$ \\
Diabetes & 10 & 1.0 & $0.0156 \pm 0.0038$ \\
Age & 10 & 1.0 & $0.0139 \pm 0.0039$ \\
GGT (log) & 10 & 1.0 & $0.0103 \pm 0.0041$ \\
Race & 10 & 1.0 & $0.0093 \pm 0.0015$ \\
Gender & 10 & 1.0 & $0.0057 \pm 0.0014$ \\
Sleep Midpoint $\times$ Age & 10 & 1.0 & $0.0020 \pm 0.0012$ \\
Androstenedione & 8 & 0.8 & $0.0035 \pm 0.0014$ \\
ALT (log) & 8 & 0.8 & $0.0033 \pm 0.0019$ \\
ACR (log) & 5 & 0.5 & $0.0017 \pm 0.0012$ \\
\bottomrule
\end{tabular}

\vspace{1mm}
\footnotesize\emph{Note:} Summary is based on 10 repeated random-split runs. Count = number of runs in which the feature was retained; Frequency = retention frequency; Importance = permutation importance.
\end{table}

\begin{table}[htbp]
\caption{Repeated random-split targeted ablation analysis for
\textit{Sleep Midpoint} $\times$ \textit{Age}.}
\label{tab:ablation_sleepmid}
\centering

\setlength{\tabcolsep}{4pt}

\begin{tabular}{@{}p{0.28\textwidth}cccc@{}}
\toprule
\textbf{Model / comparison}
& \textbf{Test $R^2$} & \textbf{MAE} & \textbf{RMSE} & \textbf{Tuned CV $R^2$} \\
\midrule

Model 5b
& $0.7459 \pm 0.0327$ & $0.08230 \pm 0.00232$ & $0.1207 \pm 0.0068$ & $0.7138 \pm 0.0096$ \\

Model 5b (ablated)
& $0.7456 \pm 0.0317$ & $0.08235 \pm 0.00250$ & $0.1208 \pm 0.0068$ & $0.7139 \pm 0.0093$ \\

\midrule

Delta (Model 5b minus ablated)
& $0.00030$ & $-0.00005$ & $-0.00009$ & $-0.00010$ \\

\bottomrule
\end{tabular}

\vspace{1mm}

\parbox{\textwidth}{%
\footnotesize
\emph{Note:} Model 5b is compared with a corresponding refitted model in which
\textit{Sleep Midpoint} $\times$ \textit{Age} was removed from the final compact
predictor set. Values are reported as mean $\pm$ standard deviation across
10 random seeds. Delta values are computed as Model 5b minus ablated model.
}

\end{table}

Across repeated random splits, predictive performance remained broadly stable for Models 4, 5, and 5b, with only small differences in mean test-set performance. The final compact model (Model 5b) preserved similar predictive accuracy while using a smaller predictor set. Several predictors were selected consistently across all runs, including GHB, diabetes diagnosis, insulin, age, race, GGT, gender, and the engineered interaction term \textit{Sleep Midpoint} $\times$ \textit{Age}. Although the mean permutation importance of \textit{Sleep Midpoint} $\times$ \textit{Age} remained small relative to dominant metabolic predictors, its 100\% selection frequency across repeated runs supports its stability within the full-model interaction pipeline. However, the targeted ablation analysis in Table~\ref{tab:ablation_sleepmid} showed that removing \textit{Sleep Midpoint} $\times$ \textit{Age} led to negligible differences in predictive performance across seeds, supporting the interpretation that this interaction was stable but modest rather than a major driver of incremental predictive gain.

\section{Restricted sensitivity analyses excluding major glycaemic predictors}\label{secA7}

To assess the extent to which the full-model results depended on dominant glycaemic predictors, restricted sensitivity analyses were performed after excluding glycohemoglobin (GHB/HbA1c), insulin, diabetes status, and closely related interaction terms from the candidate feature space. The interaction-augmented XGBoost pipeline was then repeated across 10 random seeds using the same age-stratified train-test split, tuning, SHAP-guided feature selection, pruning, and evaluation strategy as in the main repeated random-split analyses. Appendix Table~\ref{tab:restricted_performance} summarizes predictive performance across seeds, and Appendix Table~\ref{tab:restricted_feature_frequency} reports feature-selection frequency for the final restricted compact model pipeline. Notably, \textit{Sleep Midpoint} $\times$ \textit{Age} was not selected in any repeated restricted run for either restricted Model 5 or restricted Model 5b.

\begin{table*}[t]
\caption{Restricted repeated random-split robustness summary after excluding major glycaemic predictors and related interaction terms.}
\label{tab:restricted_performance}
\centering
\begin{tabular}{lcccc} 
\toprule
\textbf{Model} & \textbf{Test $R^2$} & \textbf{MAE} & \textbf{RMSE} & \textbf{Tuned CV $R^2$} \\
\midrule
Restricted Model 4 & $0.354 \pm 0.016$ & $0.1267 \pm 0.0047$ & $0.1931 \pm 0.0112$ & $0.3164 \pm 0.0071$ \\
Restricted Model 5 & $0.338 \pm 0.015$ & $0.1280 \pm 0.0046$ & $0.1955 \pm 0.0106$ & $0.3179 \pm 0.0096$ \\
Restricted Model 5b & $0.340 \pm 0.016$ & $0.1278 \pm 0.0049$ & $0.1952 \pm 0.0111$ & $0.3176 \pm 0.0093$ \\
\bottomrule
\end{tabular}

\vspace{1mm}
\footnotesize\emph{Note:} Values are reported as mean $\pm$ standard deviation across 10 random seeds.
\end{table*}

\begin{table}[t]
\caption{Feature selection frequency for the final restricted compact model (restricted Model 5b).}
\label{tab:restricted_feature_frequency}
\centering
\begin{tabular}{lcc}
\toprule
\textbf{Feature} & \textbf{Count} & \textbf{Frequency} \\
\midrule
Age & 10 & 1.0 \\
ACR (log) & 10 & 1.0 \\
GGT (log) & 10 & 1.0 \\
Triglycerides (log) & 10 & 1.0 \\
HDL & 10 & 1.0 \\
SHBG (log) & 10 & 1.0 \\
AST (log) & 10 & 1.0 \\
ALT (log) & 10 & 1.0 \\
Uric acid & 10 & 1.0 \\
BMI & 10 & 1.0 \\
eGFR & 9 & 0.9 \\
LDL & 6 & 0.6 \\
Diet & 1 & 0.1 \\
CRP (log) & 1 & 0.1 \\
Sleep hours $\times$ Age & 1 & 0.1 \\
\bottomrule
\end{tabular}

\vspace{1mm}
\footnotesize\emph{Note:} Summary is based on 10 repeated random-split runs. Count = number of runs in which the feature was retained; Frequency = retention frequency.
\end{table}

\section*{Funding}
This work did not receive any specific funding.

\section*{Acknowledgment}
The authors acknowledge the National Center for Health Statistics (NCHS) and the Centers for Disease Control and Prevention (CDC) for providing access to the publicly available NHANES 2017-2020 data.

\bibliographystyle{unsrt}
\bibliography{references}

@article{burksChronobiologicallyinformedFeaturesCGM2025,
  title = {Chronobiologically-Informed Features from {{CGM}} Data Provide Unique Information for {{XGBoost}} Prediction of Longer-Term Glycemic Dysregulation in 8,000 Individuals with Type-2 Diabetes},
  author = {Burks, Jamison H. and Joe, Leslie and Kanjaria, Karina and Monsivais, Carlos and O'laughlin, Kate and Smarr, Benjamin L.},
  year = 2025,
  month = apr,
  journal = {PLOS digital health},
  volume = {4},
  number = {4},
  pages = {e0000815},
  issn = {2767-3170},
  doi = {10.1371/journal.pdig.0000815},
  langid = {english},
  pmcid = {PMC11981153},
  pmid = {40202975}
}

@article{changApplicationMachineLearning2022,
  title = {Application of Machine Learning Methods for the Prediction of True Fasting Status in Patients Performing Blood Tests},
  author = {Chang, Shih-Ni and Hsiao, Ya-Luan and Lin, Che-Chen and Sun, Chuan-Hu and Chen, Pei-Shan and Wu, Min-Yen and Chen, Sheng-Hsuan and Chiang, Hsiu-Yin and Hsiao, Chiung-Tzu and King, Emily K. and Chang, Chun-Min and Kuo, Chin-Chi},
  year = 2022,
  month = jul,
  journal = {Scientific Reports},
  volume = {12},
  number = {1},
  pages = {11929},
  issn = {2045-2322},
  doi = {10.1038/s41598-022-15161-2},
  langid = {english},
  pmcid = {PMC9279373},
  pmid = {35831336}
}

@article{elmagarmidInvestigationRiskFactors2024,
  title = {Investigation of the Risk Factors Associated with Prediabetes in Normal-Weight {{Qatari}} Adults: A Cross-Sectional Study},
  shorttitle = {Investigation of the Risk Factors Associated with Prediabetes in Normal-Weight {{Qatari}} Adults},
  author = {Elmagarmid, Khadija A. and Fadlalla, Mohamed and Jose, Johann and Arredouani, Abdelilah and Bensmail, Halima},
  year = 2024,
  month = oct,
  journal = {Scientific Reports},
  volume = {14},
  number = {1},
  pages = {23116},
  issn = {2045-2322},
  doi = {10.1038/s41598-024-73476-8},
  langid = {english},
  pmcid = {PMC11452400},
  pmid = {39367088}
}

@article{fuStackingModelFramework2024,
  title = {Stacking Model Framework Reveals Clinical Biochemical Data and Dietary Behavior Features Associated with Type 2 Diabetes: {{A}} Retrospective Cohort Study},
  shorttitle = {Stacking Model Framework Reveals Clinical Biochemical Data and Dietary Behavior Features Associated with Type 2 Diabetes},
  author = {Fu, Yong and Liang, Xinghuan and Yang, Xi and Li, Li and Meng, Liheng and Wei, Yuekun and Huang, Daizheng and Qin, Yingfen},
  year = 2024,
  month = dec,
  journal = {APL bioengineering},
  volume = {8},
  number = {4},
  pages = {046111},
  issn = {2473-2877},
  doi = {10.1063/5.0207658},
  langid = {english},
  pmcid = {PMC11584240},
  pmid = {39583336}
}

@article{hossainMetabolicSyndromePredictive2024,
  title = {Metabolic Syndrome Predictive Modelling in {{Bangladesh}} Applying Machine Learning Approach},
  author = {Hossain, Md Farhad and Hossain, Shaheed and Akter, Mst Nira and Nahar, Ainur and Liu, Bowen and Faruque, Md Omar},
  year = 2024,
  journal = {PloS One},
  volume = {19},
  number = {9},
  pages = {e0309869},
  issn = {1932-6203},
  doi = {10.1371/journal.pone.0309869},
  langid = {english},
  pmcid = {PMC11376561},
  pmid = {39236041}
}

@article{huangIncreasingTransparencyMachine2023,
  title = {Increasing Transparency in Machine Learning through Bootstrap Simulation and Shapely Additive Explanations},
  author = {Huang, Alexander A. and Huang, Samuel Y.},
  year = 2023,
  journal = {PloS One},
  volume = {18},
  number = {2},
  pages = {e0281922},
  issn = {1932-6203},
  doi = {10.1371/journal.pone.0281922},
  langid = {english},
  pmcid = {PMC9949629},
  pmid = {36821544}
}

@article{javeedCircadianEtiologyType2018,
  title = {Circadian {{Etiology}} of {{Type}} 2 {{Diabetes Mellitus}}},
  author = {Javeed, Naureen and Matveyenko, Aleksey V.},
  year = 2018,
  month = mar,
  journal = {Physiology (Bethesda, Md.)},
  volume = {33},
  number = {2},
  pages = {138--150},
  issn = {1548-9221},
  doi = {10.1152/physiol.00003.2018},
  langid = {english},
  pmcid = {PMC5899235},
  pmid = {29412061}
}

@article{kellerRiskDiabetesLong2025,
  title = {Risk for {{Diabetes From Long Working Hours}} and {{Night Work}} in the {{United States}}: {{Prospective Associations}} and {{Machine Learning Techniques}}},
  shorttitle = {Risk for {{Diabetes From Long Working Hours}} and {{Night Work}} in the {{United States}}},
  author = {Keller, Elizabeth and Chen, Liwei and Gao, Feng and Li, Jian},
  year = 2025,
  month = sep,
  journal = {Safety and Health at Work},
  volume = {16},
  number = {3},
  pages = {355--360},
  issn = {2093-7911},
  doi = {10.1016/j.shaw.2025.05.005},
  urldate = {2026-04-02}
}

@article{kiranType2Diabetes2026,
  title = {Type 2 Diabetes Prediction without Labs: A Systems-Level Neural Framework for Risk and Behavioral Network Reorganization},
  shorttitle = {Type 2 Diabetes Prediction without Labs},
  author = {Kiran, Mahreen and Xie, Ying and Ball, Graham and Anjum, Nasreen and Schutte, Rudolph and Pierscionek, Barbara},
  year = 2026,
  month = jan,
  journal = {Frontiers in Digital Health},
  volume = {7},
  publisher = {Frontiers},
  issn = {2673-253X},
  doi = {10.3389/fdgth.2025.1714545},
  urldate = {2026-04-02},
  langid = {english}
}

@article{krauseType1Type2023,
  title = {Type 1 and {{Type}} 2 {{Diabetes Mellitus}}: {{Commonalities}}, {{Differences}} and the {{Importance}} of {{Exercise}} and {{Nutrition}}},
  shorttitle = {Type 1 and {{Type}} 2 {{Diabetes Mellitus}}},
  author = {Krause, Maur{\'i}cio and De Vito, Giuseppe},
  year = 2023,
  month = oct,
  journal = {Nutrients},
  volume = {15},
  number = {19},
  pages = {4279},
  issn = {2072-6643},
  doi = {10.3390/nu15194279},
  langid = {english},
  pmcid = {PMC10574155},
  pmid = {37836562}
}

@article{laiPredictiveModelsDiabetes2019,
  title = {Predictive Models for Diabetes Mellitus Using Machine Learning Techniques},
  author = {Lai, Hang and Huang, Huaxiong and Keshavjee, Karim and Guergachi, Aziz and Gao, Xin},
  year = 2019,
  month = oct,
  journal = {BMC endocrine disorders},
  volume = {19},
  number = {1},
  pages = {101},
  issn = {1472-6823},
  doi = {10.1186/s12902-019-0436-6},
  langid = {english},
  pmcid = {PMC6794897},
  pmid = {31615566}
}

@article{liMachineLearningPredicting2023,
  title = {Machine Learning for Predicting Diabetes Risk in Western {{China}} Adults},
  author = {Li, Lin and Cheng, Yinlin and Ji, Weidong and Liu, Mimi and Hu, Zhensheng and Yang, Yining and Wang, Yushan and Zhou, Yi},
  year = 2023,
  month = jul,
  journal = {Diabetology \& Metabolic Syndrome},
  volume = {15},
  number = {1},
  pages = {165},
  issn = {1758-5996},
  doi = {10.1186/s13098-023-01112-y},
  langid = {english},
  pmcid = {PMC10373320},
  pmid = {37501094}
}

@article{liuMachineLearningModels2023,
  title = {Machine {{Learning Models}} for {{Blood Glucose Level Prediction}} in {{Patients With Diabetes Mellitus}}: {{Systematic Review}} and {{Network Meta-Analysis}}},
  shorttitle = {Machine {{Learning Models}} for {{Blood Glucose Level Prediction}} in {{Patients With Diabetes Mellitus}}},
  author = {Liu, Kui and Li, Linyi and Ma, Yifei and Jiang, Jun and Liu, Zhenhua and Ye, Zichen and Liu, Shuang and Pu, Chen and Chen, Changsheng and Wan, Yi},
  year = 2023,
  month = nov,
  journal = {JMIR medical informatics},
  volume = {11},
  pages = {e47833},
  issn = {2291-9694},
  doi = {10.2196/47833},
  langid = {english},
  pmcid = {PMC10696506},
  pmid = {37983072}
}

@article{liuUseMachineLearning2024,
  title = {Use of {{Machine Learning}} to {{Predict}} the {{Incidence}} of {{Type}} 2 {{Diabetes Among Relatively Healthy Adults}}: {{A}} 10-{{Year Longitudinal Study}} in {{Taiwan}}},
  shorttitle = {Use of {{Machine Learning}} to {{Predict}} the {{Incidence}} of {{Type}} 2 {{Diabetes Among Relatively Healthy Adults}}},
  author = {Liu, Ying-Qiang and Chang, Tzu-Wei and Lee, Lung-Chun and Chen, Chia-Yu and Hsu, Pi-Shan and Tsan, Yu-Tse and Yang, Chao-Tung and Chu, Wei-Min},
  year = 2024,
  month = dec,
  journal = {Diagnostics (Basel, Switzerland)},
  volume = {15},
  number = {1},
  pages = {72},
  issn = {2075-4418},
  doi = {10.3390/diagnostics15010072},
  langid = {english},
  pmcid = {PMC11719639},
  pmid = {39795600}
}

@article{lvDetectionDiabeticPatients2023,
  title = {Detection of Diabetic Patients in People with Normal Fasting Glucose Using Machine Learning},
  author = {Lv, Kun and Cui, Chunmei and Fan, Rui and Zha, Xiaojuan and Wang, Pengyu and Zhang, Jun and Zhang, Lina and Ke, Jing and Zhao, Dong and Cui, Qinghua and Yang, Liming},
  year = 2023,
  month = sep,
  journal = {BMC medicine},
  volume = {21},
  number = {1},
  pages = {342},
  issn = {1741-7015},
  doi = {10.1186/s12916-023-03045-9},
  langid = {english},
  pmcid = {PMC10483877},
  pmid = {37674168}
}

@article{maghsoudipourAssociationsChronotypeSleep2022,
  title = {Associations of Chronotype and Sleep Patterns with Metabolic Syndrome in the {{Hispanic}} Community Health Study/Study of {{Latinos}}},
  author = {Maghsoudipour, Maryam and Allison, Matthew A. and Patel, Sanjay R. and Talavera, Gregory A. and Daviglus, Martha and Zee, Phyllis C. and Reid, Kathryn J. and Makarem, Nour and Malhotra, Atul},
  year = 2022,
  month = aug,
  journal = {Chronobiology International},
  volume = {39},
  number = {8},
  pages = {1087--1099},
  issn = {1525-6073},
  doi = {10.1080/07420528.2022.2069030},
  langid = {english},
  pmcid = {PMC9177706},
  pmid = {35509113}
}

@article{manoogianCircadianClockNutrient2016,
  title = {Circadian Clock, Nutrient Quality, and Eating Pattern Tune Diurnal Rhythms in the Mitochondrial Proteome},
  author = {Manoogian, Emily N. C. and Panda, Satchidananda},
  year = 2016,
  month = mar,
  journal = {Proceedings of the National Academy of Sciences of the United States of America},
  volume = {113},
  number = {12},
  pages = {3127--3129},
  issn = {1091-6490},
  doi = {10.1073/pnas.1601786113},
  langid = {english},
  pmcid = {PMC4812772},
  pmid = {26979954}
}

@article{morrisEffectsInternalCircadian2016,
  title = {Effects of the {{Internal Circadian System}} and {{Circadian Misalignment}} on {{Glucose Tolerance}} in {{Chronic Shift Workers}}},
  author = {Morris, Christopher J. and Purvis, Taylor E. and Mistretta, Joseph and Scheer, Frank A. J. L.},
  year = 2016,
  month = mar,
  journal = {The Journal of Clinical Endocrinology and Metabolism},
  volume = {101},
  number = {3},
  pages = {1066--1074},
  issn = {1945-7197},
  doi = {10.1210/jc.2015-3924},
  langid = {english},
  pmcid = {PMC4803172},
  pmid = {26771705}
}

@article{morrisEndogenousCircadianSystem2015,
  title = {Endogenous Circadian System and Circadian Misalignment Impact Glucose Tolerance via Separate Mechanisms in Humans},
  author = {Morris, Christopher J. and Yang, Jessica N. and Garcia, Joanna I. and Myers, Samantha and Bozzi, Isadora and Wang, Wei and Buxton, Orfeu M. and Shea, Steven A. and Scheer, Frank A. J. L.},
  year = 2015,
  month = apr,
  journal = {Proceedings of the National Academy of Sciences of the United States of America},
  volume = {112},
  number = {17},
  pages = {E2225-2234},
  issn = {1091-6490},
  doi = {10.1073/pnas.1418955112},
  langid = {english},
  pmcid = {PMC4418873},
  pmid = {25870289}
}

@article{nortonInsulinMasterRegulator2022,
  title = {Insulin: {{The}} Master Regulator of Glucose Metabolism},
  shorttitle = {Insulin},
  author = {Norton, Luke and Shannon, Chris and Gastaldelli, Amalia and DeFronzo, Ralph A.},
  year = 2022,
  month = apr,
  journal = {Metabolism: Clinical and Experimental},
  volume = {129},
  pages = {155142},
  issn = {1532-8600},
  doi = {10.1016/j.metabol.2022.155142},
  langid = {english},
  pmid = {35066003}
}

@article{opperhuizenLightNightAcutely2017,
  title = {Light at Night Acutely Impairs Glucose Tolerance in a Time-, Intensity- and Wavelength-Dependent Manner in Rats},
  author = {Opperhuizen, Anne-Loes and Stenvers, Dirk J. and Jansen, Remi D. and Foppen, Ewout and Fliers, Eric and Kalsbeek, Andries},
  year = 2017,
  month = jul,
  journal = {Diabetologia},
  volume = {60},
  number = {7},
  pages = {1333--1343},
  issn = {1432-0428},
  doi = {10.1007/s00125-017-4262-y},
  langid = {english},
  pmcid = {PMC5487588},
  pmid = {28374068}
}

@article{phillipsUncoveringPersonalizedGlucose2023,
  title = {Uncovering Personalized Glucose Responses and Circadian Rhythms from Multiple Wearable Biosensors with {{Bayesian}} Dynamical Modeling},
  author = {Phillips, Nicholas E. and Collet, Tinh-Hai and Naef, Felix},
  year = 2023,
  month = aug,
  journal = {Cell Reports Methods},
  volume = {3},
  number = {8},
  pages = {100545},
  issn = {2667-2375},
  doi = {10.1016/j.crmeth.2023.100545},
  langid = {english},
  pmcid = {PMC10475794},
  pmid = {37671030}
}

@article{rothCircadianmediatedRegulationCardiometabolic2023,
  title = {Circadian-Mediated Regulation of Cardiometabolic Disorders and Aging with Time-Restricted Feeding},
  author = {Roth, Jonathan R. and Varshney, Shweta and {de Moraes}, Ruan Carlos Macedo and Melkani, Girish C.},
  year = 2023,
  month = feb,
  journal = {Obesity (Silver Spring, Md.)},
  volume = {31 Suppl 1},
  number = {Suppl 1},
  pages = {40--49},
  issn = {1930-739X},
  doi = {10.1002/oby.23664},
  langid = {english},
  pmcid = {PMC10089654},
  pmid = {36623845}
}

@article{sadriaAgingAffectsCircadian2021,
  title = {Aging Affects Circadian Clock and Metabolism and Modulates Timing of Medication},
  author = {Sadria, Mehrshad and Layton, Anita T.},
  year = 2021,
  month = apr,
  journal = {iScience},
  volume = {24},
  number = {4},
  pages = {102245},
  issn = {2589-0042},
  doi = {10.1016/j.isci.2021.102245},
  langid = {english},
  pmcid = {PMC7995490},
  pmid = {33796837}
}

@article{shenTrajectoriesSleepDuration2025,
  title = {Trajectories of {{Sleep Duration}}, {{Sleep Onset Timing}}, and {{Continuous Glucose Monitoring}} in {{Adults}}},
  author = {Shen, Luqi and Li, Bang-Yan and Gou, Wanglong and Liang, Xinxiu and Zhong, Haili and Xiao, Congmei and Shi, Ruiqi and Miao, Zelei and Yan, Yan and Fu, Yuanqing and Chen, Yu-Ming and Zheng, Ju-Sheng},
  year = 2025,
  month = mar,
  journal = {JAMA network open},
  volume = {8},
  number = {3},
  pages = {e250114},
  issn = {2574-3805},
  doi = {10.1001/jamanetworkopen.2025.0114},
  langid = {english},
  pmcid = {PMC11883496},
  pmid = {40042843}
}

@article{shojaee-mendPredictionDiabetesUsing2024,
  title = {Prediction of {{Diabetes Using Data Mining}} and {{Machine Learning Algorithms}}: {{A Cross-Sectional Study}}},
  shorttitle = {Prediction of {{Diabetes Using Data Mining}} and {{Machine Learning Algorithms}}},
  author = {{Shojaee-Mend}, Hassan and Velayati, Farnia and Tayefi, Batool and Babaee, Ebrahim},
  year = 2024,
  month = jan,
  journal = {Healthcare Informatics Research},
  volume = {30},
  number = {1},
  pages = {73--82},
  issn = {2093-3681},
  doi = {10.4258/hir.2024.30.1.73},
  langid = {english},
  pmcid = {PMC10879823},
  pmid = {38359851}
}

@article{taoPredictingThreemonthFasting2023,
  title = {Predicting Three-Month Fasting Blood Glucose and Glycated Hemoglobin Changes in Patients with Type 2 Diabetes Mellitus Based on Multiple Machine Learning Algorithms},
  author = {Tao, Xue and Jiang, Min and Liu, Yumeng and Hu, Qi and Zhu, Baoqiang and Hu, Jiaqiang and Guo, Wenmei and Wu, Xingwei and Xiong, Yu and Shi, Xia and Zhang, Xueli and Han, Xu and Li, Wenyuan and Tong, Rongsheng and Long, Enwu},
  year = 2023,
  month = sep,
  journal = {Scientific Reports},
  volume = {13},
  number = {1},
  pages = {16437},
  issn = {2045-2322},
  doi = {10.1038/s41598-023-43240-5},
  langid = {english},
  pmcid = {PMC10543442},
  pmid = {37777593}
}

@article{tengAssociationSerumGamma2023,
  title = {Association between Serum Gamma Glutamyl Transferase and Fasting Blood Glucose in {{Chinese}} People: {{A}} 6-Year Follow-up Study},
  shorttitle = {Association between Serum Gamma Glutamyl Transferase and Fasting Blood Glucose in {{Chinese}} People},
  author = {Teng, Fei and Ye, Yan and Wang, Liying and Qin, Ruihao and Liu, Xuekui and Geng, Houfa and Xu, Wei and Lai, Peng and Liang, Jun},
  year = 2023,
  month = feb,
  journal = {Journal of Diabetes Investigation},
  volume = {14},
  number = {2},
  pages = {339--343},
  issn = {2040-1124},
  doi = {10.1111/jdi.13947},
  langid = {english},
  pmcid = {PMC9889614},
  pmid = {36412546}
}

@article{tranEffectCircadianClock2024,
  title = {Effect of Circadian Clock Disruption on Type 2 Diabetes},
  author = {Tran, Hong Thuan and Kondo, Takeru and Ashry, Amal and Fu, Yunyu and Okawa, Hiroko and Sawangmake, Chenphop and Egusa, Hiroshi},
  year = 2024,
  journal = {Frontiers in Physiology},
  volume = {15},
  pages = {1435848},
  issn = {1664-042X},
  doi = {10.3389/fphys.2024.1435848},
  langid = {english},
  pmcid = {PMC11333352},
  pmid = {39165284}
}

@article{vandoornMachineLearningbasedGlucose2021,
  title = {Machine Learning-Based Glucose Prediction with Use of Continuous Glucose and Physical Activity Monitoring Data: {{The Maastricht Study}}},
  shorttitle = {Machine Learning-Based Glucose Prediction with Use of Continuous Glucose and Physical Activity Monitoring Data},
  author = {{van Doorn}, William P. T. M. and Foreman, Yuri D. and Schaper, Nicolaas C. and Savelberg, Hans H. C. M. and Koster, Annemarie and {van der Kallen}, Carla J. H. and Wesselius, Anke and Schram, Miranda T. and Henry, Ronald M. A. and Dagnelie, Pieter C. and {de Galan}, Bastiaan E. and Bekers, Otto and Stehouwer, Coen D. A. and Meex, Steven J. R. and Brouwers, Martijn C. G. J.},
  year = 2021,
  journal = {PloS One},
  volume = {16},
  number = {6},
  pages = {e0253125},
  issn = {1932-6203},
  doi = {10.1371/journal.pone.0253125},
  langid = {english},
  pmcid = {PMC8224858},
  pmid = {34166426}
}

@article{wollerCircadianMisalignmentMetabolic2021,
  title = {Circadian {{Misalignment}} and {{Metabolic Disorders}}: {{A Story}} of {{Twisted Clocks}}},
  shorttitle = {Circadian {{Misalignment}} and {{Metabolic Disorders}}},
  author = {Woller, Aurore and Gonze, Didier},
  year = 2021,
  month = mar,
  journal = {Biology},
  volume = {10},
  number = {3},
  pages = {207},
  issn = {2079-7737},
  doi = {10.3390/biology10030207},
  langid = {english},
  pmcid = {PMC8001388},
  pmid = {33801795}
}

@article{wuRiskFactorsContributing2014,
  title = {Risk Factors Contributing to Type 2 Diabetes and Recent Advances in the Treatment and Prevention},
  author = {Wu, Yanling and Ding, Yanping and Tanaka, Yoshimasa and Zhang, Wen},
  year = 2014,
  journal = {International Journal of Medical Sciences},
  volume = {11},
  number = {11},
  pages = {1185--1200},
  issn = {1449-1907},
  doi = {10.7150/ijms.10001},
  langid = {english},
  pmcid = {PMC4166864},
  pmid = {25249787}
}

@article{xuRestactivityCircadianRhythm2022,
  title = {Rest-Activity Circadian Rhythm and Impaired Glucose Tolerance in Adults: An Analysis of {{NHANES}} 2011-2014},
  shorttitle = {Rest-Activity Circadian Rhythm and Impaired Glucose Tolerance in Adults},
  author = {Xu, Yanyan and Su, Shaoyong and McCall, William V. and Isales, Carlos and Snieder, Harold and Wang, Xiaoling},
  year = 2022,
  month = mar,
  journal = {BMJ open diabetes research \& care},
  volume = {10},
  number = {2},
  pages = {e002632},
  issn = {2052-4897},
  doi = {10.1136/bmjdrc-2021-002632},
  langid = {english},
  pmcid = {PMC8895931},
  pmid = {35241430}
}

@article{yuEveningChronotypeAssociated2015,
  title = {Evening Chronotype Is Associated with Metabolic Disorders and Body Composition in Middle-Aged Adults},
  author = {Yu, Ji Hee and Yun, Chang-Ho and Ahn, Jae Hee and Suh, Sooyeon and Cho, Hyun Joo and Lee, Seung Ku and Yoo, Hye Jin and Seo, Ji A. and Kim, Sin Gon and Choi, Kyung Mook and Baik, Sei Hyun and Choi, Dong Seop and Shin, Chol and Kim, Nan Hee},
  year = 2015,
  month = apr,
  journal = {The Journal of Clinical Endocrinology and Metabolism},
  volume = {100},
  number = {4},
  pages = {1494--1502},
  issn = {1945-7197},
  doi = {10.1210/jc.2014-3754},
  langid = {english},
  pmid = {25831477}
}

@article{zhangCircadianRhythmGlucose2025,
  title = {Circadian Rhythm, Glucose Metabolism and Diabetic Complications: The Role of Glucokinase and the Enlightenment on Future Treatment},
  shorttitle = {Circadian Rhythm, Glucose Metabolism and Diabetic Complications},
  author = {Zhang, Zhijun and Wang, Shuo and Gao, Ling},
  year = 2025,
  journal = {Frontiers in Physiology},
  volume = {16},
  pages = {1537231},
  issn = {1664-042X},
  doi = {10.3389/fphys.2025.1537231},
  langid = {english},
  pmcid = {PMC11885239},
  pmid = {40061454}
}

@article{zhangDevelopmentValidationMachine2024,
  title = {Development and {{Validation}} of {{Machine Learning Models}} for {{Identifying Prediabetes}} and {{Diabetes}} in {{Normoglycemia}}},
  author = {Zhang, Xiaodong and Yao, Weidong and Wang, Dawei and Hu, Wenqi and Zhang, Guang and Zhang, Yongsheng},
  year = 2024,
  month = nov,
  journal = {Diabetes/Metabolism Research and Reviews},
  volume = {40},
  number = {8},
  pages = {e70003},
  issn = {1520-7560},
  doi = {10.1002/dmrr.70003},
  langid = {english},
  pmcid = {PMC11601146},
  pmid = {39497474}
}

@article{zhangEarlyDetectionType2023,
  title = {Early Detection of Type 2 Diabetes Risk: Limitations of Current Diagnostic Criteria},
  shorttitle = {Early Detection of Type 2 Diabetes Risk},
  author = {Zhang, Jiale and Zhang, Zhuoya and Zhang, Kaiqi and Ge, Xiaolei and Sun, Ranran and Zhai, Xu},
  year = 2023,
  journal = {Frontiers in Endocrinology},
  volume = {14},
  pages = {1260623},
  issn = {1664-2392},
  doi = {10.3389/fendo.2023.1260623},
  langid = {english},
  pmcid = {PMC10665905},
  pmid = {38027114}
}

\end{document}